\documentclass{ecai} 

\makeatletter
\let\tufte@caption\@caption  
\usepackage{hyperref}
\let\@caption\tufte@caption  

\usepackage{latexsym}
\usepackage[table]{xcolor}
\usepackage{amssymb}
\usepackage{amsmath}
\usepackage{amsthm}
\usepackage{siunitx}
\usepackage{mathtools}
\usepackage{booktabs}
\usepackage{enumitem}
\usepackage{graphicx}
\usepackage{color}
\usepackage{tikz}
\usepackage{url}
\usepackage{pifont}
\usepackage[many]{tcolorbox}
\usepackage{multirow}
\usepackage{anyfontsize}
\usepackage{tabularx}
\usepackage{xfrac}

\usepackage[frozencache=true,cachedir=minted-cache]{minted}
\usetikzlibrary{positioning, fadings, calc, shapes.callouts,shapes.arrows, bayesnet, fit, backgrounds}

\newcommand{\BibTeX}{B\kern-.05em{\sc i\kern-.025em b}\kern-.08em\TeX}
\definecolor{color_annot_green}{RGB}{0,128,128}
\definecolor{color_annot_blue}{RGB}{42,127,255}
\definecolor{color_annot_violet}{RGB}{127,0,127}
\definecolor{pink_ground_truth}{RGB}{128,0,128}
\definecolor{green_annotations}{RGB}{0,128,128}
\definecolor{datasetcolor}{rgb}{0.302,0.302,0.302}
\newcommand{\best}[1]{{\color{blue}{\textBF{#1}}}}
\newcommand{\secondbest}[1]{{\color{purple}{\textBF{#1}}}}
\newcommand{\excluded}[1]{{\color{orange}{#1}}}
\newcommand{\worse}[1]{{\underline{#1}}}
\newcommand{\defas}{\coloneqq}
\newcommand{\cond}{\,|\,}
\DeclareMathOperator*{\argmax}{arg\,max}

\newsavebox\CBox
\def\textBF#1{\sbox\CBox{#1}\resizebox{\wd\CBox}{\ht\CBox}{\textbf{#1}}}
\newcounter{loopc}
\NewDocumentCommand\towrite{O{1}m}%
  {{\color{red}#2\forloop{loopc}{1}{\value{loopc} < #1}{; #2}}}
\newdimen\abovecrulesep
\newdimen\belowcrulesep
\makeatletter
\patchcmd{\@@@cmidrule}{\aboverulesep}{\abovecrulesep}{}{}
\patchcmd{\@xcmidrule}{\belowrulesep}{\belowcrulesep}{}{}
\makeatother
\makeatletter
\@ifpackageloaded{hyperref}{
  \newcommand{\github}{\url{https://github.com/ies-research/multi-annotator-machine-learning/tree/annot-mix}}
  \newcommand{\mymail}{\href{mailto:marek.herde@uni-kassel.de}{marek.herde@uni-kassel.de}}
}{
  \newcommand{\github}{https://github.com/ies-research/multi-annotator-machine-learning/tree/annot-mix}
  \newcommand{\mymail}{marek.herde@uni-kassel.de}
}
\makeatother
\makeatletter
\newcommand{\footnoteref}[1]{%
	\ltx@ifpackageloaded{hyperref}{
		\ifHy@hyperfootnotes
		\hbox{\hyperref[#1]{%
				\@textsuperscript {\normalfont \ref*{#1}}}}%
		\else
		\hbox{\@textsuperscript {\normalfont \ref*{#1}}}%
		\fi%
	}{
		\hbox{\@textsuperscript {\normalfont \ref{#1}}}%
	}%
}
\makeatother

\begin{document}


\begin{frontmatter}


\paperid{2297} 


\title{%
{\footnotesize
    \vspace{-5em}
    {\footnotesize
    \begin{center}
    \begin{minipage}{0.93\textwidth}
        \centering
        \copyright\ Marek Herde, Lukas Lührs, Denis Huseljic, and Bernhard Sick, 2024.
        The definitive, peer-reviewed, and edited version of this article is published in
        \textit{ECAI 2024: 27th European Conference on Artificial Intelligence,
        19--24 October 2024, Santiago de Compostela, Spain -- Including 13th Conference
        on Prestigious Applications of Intelligent Systems (PAIS 2024)},
        edited by U. Endriss, F. S. Melo, K. Bach, A. Bugarín-Diz, J. M. Alonso-Moral,
        S. Barro, and F. Heintz, IOS Press, ISBN online 978-1-64368-548-9,
        pp.~2910--2918, 2024.
        DOI: \href{https://doi.org/10.3233/FAIA240829}{10.3233/FAIA240829}. This arXiv version includes editorial corrections that do not affect the methodology, results, or conclusions.
    \end{minipage}
    \end{center}
    }
    \vspace{1.5em}
}

Annot-Mix: Learning with Noisy Class Labels from Multiple Annotators via a Mixup Extension
}

\author[A]{\fnms{Marek}~\snm{Herde}\thanks{Corresponding Author. Email: \mymail.}}
\author[A]{\fnms{Lukas}~\snm{Lührs}}
\author[A]{\fnms{Denis}~\snm{Huseljic}} 
\author[A]{\fnms{Bernhard}~\snm{Sick}}

\address[A]{University of Kassel, Germany}

\begin{abstract}
Training with noisy class labels impairs neural networks' generalization performance. In this context, \texttt{mixup} is a popular regularization technique to improve training robustness by making memorizing false class labels more difficult. However, \texttt{mixup} neglects that multiple annotators, e.g., crowdworkers, typically provide class labels. Therefore, we propose an extension of \texttt{mixup}, which handles multiple class labels per instance while considering which class label originates from which annotator. Integrated into our multi-annotator classification framework \texttt{annot-mix}, it performs superiorly to eleven (mostly state-of-the-art) approaches in an evaluation study with eleven datasets comprising noisy class labels from either human or simulated annotators. Our code is publicly available through our GitHub repository at~\github.
\end{abstract}

\end{frontmatter}


\section{Introduction}
\label{sec:introduction}
\begin{figure}[!t]
    \centering
    \begin{tikzpicture}[scale=0.92]
        \draw[-] (-0.0, 0.95) -- (9, 0.95) node[right] {};
        \draw[-] (-0.0, 3) -- (1.05, 3) node[above=0.05] {\scriptsize{\textBF{Instances}}} -- (2.95, 3) node[above=-0.05, align=center] (classification) {\scriptsize{\textBF{Classification}}} -- (6.75, 3) node[above=0.05]{\scriptsize{\textBF{Multi-annotator Classification}}} -- (9, 3);
        \node[above=-0.15 of classification]{\scriptsize{\textBF{Standard}}};
        \draw[-] (0.3, -0.15) -- (0.3, 0.425) node[above, xshift=1.5, rotate=90] {\scriptsize{\textBF{Output}}} -- (0.3, 2.0) node[above, xshift=1.5, rotate=90] {\scriptsize{\textBF{Input}}} -- (0.3, 3.5);
        \draw[-] (1.8, -0.15) -- (1.8, 3.5) node[right] {};
        \draw[-] (4.1, -0.15) -- (4.1, 3.5) node[right] {};
        \node (annot_1_pos_1)  at (4.5, 2.5) {\includegraphics[height=.7 cm]{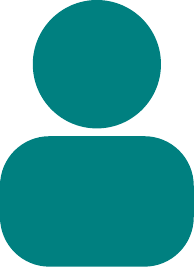}};
        \node (annot_2_pos_1) at ($(annot_1_pos_1)+(2.4,0.0)$) {\includegraphics[height=.7 cm]{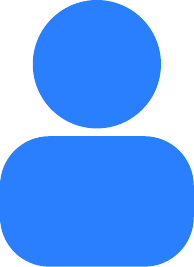}};
        \node[overlay,ellipse callout, callout relative pointer={(-0.4,0.1)}, fill=color_annot_green!50] 
            at ($(annot_1_pos_1)+(1.25,-0.1)$) {\scriptsize Jaguar};
        \node[overlay,ellipse callout, callout relative pointer={(-0.35,0.1)}, fill=color_annot_blue!50] 
             at ($(annot_2_pos_1)+(1.25,-0.1)$) {\scriptsize Leopard};
        \node[align=center] at (6.6, 0.35) {
             \begin{tcolorbox}[width=4.42cm, left=0.0cm, right=0.0cm, boxrule=0.5pt, top=0.0cm, bottom=0.0cm] \centering \scriptsize{\textBF{Research question:} How can \texttt{mixup} be effectively extended to multi-annotator classification tasks?} \end{tcolorbox}};
   
        \node (annot_1_pos_2) at ($(annot_1_pos_1)+(0.0,-1.0)$) {\includegraphics[height=.7 cm]{figures/Drawing_green_annotator.pdf}};
        \node (annot_2_pos_2) at ($(annot_2_pos_1)+(0.0,-1.0)$) {\includegraphics[height=.7 cm]{figures/Drawing_blue_annotator.pdf}};
        \node[overlay,ellipse callout, callout relative pointer={(-0.35,0.1)}, fill=color_annot_green!50] 
            at ($(annot_1_pos_2)+(1.25,-0.1)$) {\scriptsize Leopard};
        \node[] 
            at ($(annot_2_pos_2)+(1.25,-0.1)$) {\scriptsize N/A};
   
        \node (label_1) at (2.95, 2.5) {\scriptsize Jaguar};
        \node (label_2) at ($(label_1)+(0.0,-1.0)$) {\scriptsize Leopard};
        \node[align=center] (label_mix_up) at (2.95, 0.4) {\scriptsize Leopard \si{50}{\%} \\ \scriptsize Jaguar \si{50}{\%} };
   
        \node at (1.05, 2.46) {\includegraphics[height=0.8cm]{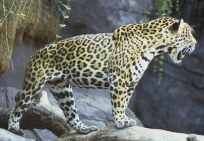}};
        \node at (1.05, 1.5) {\includegraphics[height=0.8cm]{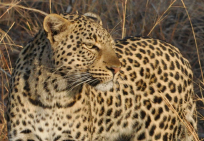}};
        \node at (1.05, 0.4) {\includegraphics[height=0.8cm]{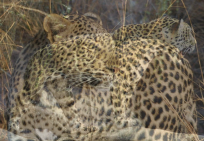}};
\end{tikzpicture}
\caption{Illustration of vanilla \texttt{mixup} and our research question: In standard classification tasks  (cf. first and second column), \texttt{mixup} convexly combines the two animal images (cf. acknowledgments for crediting USFWS) and their class labels. In contrast, multi-annotator classification tasks (cf. first and third column) allow multiple class labels to be assigned to a single instance. Further, we know which class label originates from which annotator, and some class labels may not be available (N/A) from some annotators. Hence, we must extend \texttt{mixup} toward such tasks.}
\label{fig:abstract}
\end{figure}

Training machine learning models, such as deep neural networks (DNNs), to solve classification tasks requires data instances with associated class labels, typically acquired from human annotators, e.g., crowdworkers~\cite{vaughan2017making}, in a labor-intensive process. Such annotators may be prone to errors for various reasons, e.g., lack of domain expertise, exhaustion, or disinterest~\cite{herde2021survey}. The resulting annotation errors, called noisy class labels~\cite{frenay2013classification}, impair NNs' generalization performance because NNs easily overfit on training data by memorizing noisy class labels~\cite{song2022learning}. Consequently, various approaches have been proposed to address this issue. A popular data augmentation and regularization technique is \texttt{mixup}~\cite{zhang2018mixup}, whose idea is to generate convex combinations of pairs of instances and their respective class labels (cf.\ the first and second column of~Fig.~\ref{fig:abstract} for an example in a standard classification setting). This widely applicable augmentation during the training of NNs makes pure memorization more difficult and thus reduces sensitivity to class label noise. Despite its simplicity and effectiveness, \texttt{mixup} has not been fully extended to classification tasks with class labels provided by multiple annotators, often referred to as multi-annotator classification~\cite{herde2023multi} or learning from crowds~\cite{raykar2010learning}. In this context, we face two major challenges:
\begin{itemize}
    \item \texttt{mixup} ignores that multiple class labels from varying numbers of annotators can be assigned to a single data instance.
    \item \texttt{mixup} ignores which class label originates from which annotator.
\end{itemize}
Motivated by these challenges, Fig.~\ref{fig:abstract} formulates our central research question, which we address through the following contributions:
\begin{itemize}
    \item We propose a \texttt{mixup} extension that handles multiple class labels per instance and considers each label's annotator.
    \item We integrate this extension into our multi-annotator classification approach \texttt{annot-mix}, which estimates each annotator's performance while training an NN as a classification model.
    \item We present an extensive experimental evaluation study demonstrating the superior performance of $\texttt{annot-mix}$ compared to eleven mostly state-of-the-art approaches across image, text, and tabular datasets with either human or simulated noisy class labels.
\end{itemize}

This article's remainder is structured as follows: Section~\ref{sec:problem-setup} formally introduces the problem setup of multi-annotator classification tasks. Subsequently, we discuss related work of multi-annotator classification and existing variants of \texttt{mixup} in Section~\ref{sec:related-work}. Section~\ref{sec:approach} presents our approach \texttt{annot-mix}, which is evaluated, including an ablation and hyperparameter study, in Section~\ref{sec:evaluation}. We conclude this work with an outlook on future research in Section~\ref{sec:conclusion}.

\section{Problem Setup}
\label{sec:problem-setup}
\begin{figure}[!t]
    \centering
    \begin{tikzpicture}

        \node[obs] (z) {$\boldsymbol{z}_{nm}$};
        \node[latent, left=2.5 of z] (y) {$\boldsymbol{y}_n$};
        \node[obs, above=0.5 of y] (x) {$\boldsymbol{x}_n$};
        \node[obs, above right =0.9 and =0.85 of z] (a) {$\boldsymbol{a}_m$};

        \node[left= 0cm of x] (jaguar) {\includegraphics[height=1.1cm]{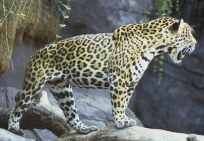}};
        \node[obs, above=0.5 of y] {$\boldsymbol{x}_n$};
        
        \node[right= 0cm of a] (green_annot){\includegraphics[height=1.1 cm]{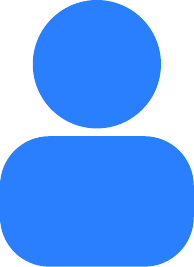}};

        \node[left=  0cm of y]{\scriptsize Jaguar};
  
        \edge {x, y, a} {z}; 
        \edge {x} {y};

        \plate {az} {(a)(z)(green_annot)} {$M$};
        \plate {yx} {(jaguar)(x)(y)(z)(az.north west)(az.south west)} {$N$};

        \node[below right=0cm of z, xshift=0.2cm, overlay,ellipse callout, callout relative pointer={(-0.2,0.1)}, fill=color_annot_blue!50] (speech_bubble) {\scriptsize Leopard};
        
    \end{tikzpicture}
    \caption{Probabilistic graphical model of the data generation in multi-annotator classification: Arrows indicate dependencies between random variables, shaded circles observable random variables, and white circles latent random variables.}
    \label{fig:probabilistic-graphical-model}
\end{figure}

Figure~\ref{fig:probabilistic-graphical-model} depicts the probabilistic graphical model that overviews the random variables and their dependencies of the commonly assumed data generation process in multi-annotator classification ~\cite{herde2023multi,li2022beyond}. More concretely, there is a multi-set $\mathcal{X} \defas \{\boldsymbol{x}_n\}_{n=1}^N \subset \Omega_X \defas \mathbb{R}^D$ of $N \in \mathbb{N}_{>0}$ instances as $D \in \mathbb{N}_{>0}$-dimensional vectors, which are independently sampled from the distribution $\Pr(\boldsymbol{x})$. Their $C \in \mathbb{N}_{>1}$-dimensional one-hot encoded true class labels form a multi-set $\mathcal{Y} \defas \{\boldsymbol{y}_n\}_{n=1}^{N} \subseteq \Omega_Y \defas \{\boldsymbol{e}_c\}_{c=1}^C$, where $C$ denotes the number of classes. In standard classification, a class label $\boldsymbol{y}_n$ is observed and sampled from the categorical distribution $\Pr(\boldsymbol{y} \cond \boldsymbol{x}_n)$. However, in multi-annotator classification, we do not know the true class labels~$\mathcal{Y}$. Instead, $M \in \mathbb{N}_{>1}$ error-prone annotators, denoted as a multi-set of vectors $\mathcal{A} \defas \{\boldsymbol{a}_m\}_{m=1}^M \subseteq \Omega_A$, independently provide noisy class labels. Throughout this article, we identify each annotator $\boldsymbol{a}_m$ by an $M$-dimensional one-hot encoded vector $\boldsymbol{e}^\prime_m$ such that $\Omega_A \defas \{\boldsymbol{e}^\prime_1, \dots, \boldsymbol{e}^\prime_M\}$. In principle, other representations would also be conceivable if metadata about the annotators~\cite{zhang2023learning} were available, e.g., annotators' levels of education. The noisy class labels are denoted as the multi-set $\mathcal{Z} \defas \{\boldsymbol{z}_{nm}\}_{n=1, m=1}^{N, M} \subseteq \Omega_\mathcal{Z} \defas \Omega_Y \cup \{\boldsymbol{0}\}$. Thereby, $\boldsymbol{z}_{nm} \in \Omega_Y$, sampled from the categorical distribution $\Pr(\boldsymbol{z} \cond \boldsymbol{x}_n, \boldsymbol{y}_n, \boldsymbol{a}_m)$, is the class label assigned by annotator $\boldsymbol{a}_m$ to instance $\boldsymbol{x}_n$ with the true class label~$\boldsymbol{y}_n$. In the case of $\boldsymbol{z}_{nm} = \boldsymbol{0}$, the annotator $\boldsymbol{a}_m$ has not annotated instance~$\boldsymbol{x}_n$, e.g., due to a limited annotation budget~\cite{khetan2018learning}.

Based on the above setup, the objective in multi-annotator classification tasks is as follows:
\begin{tcolorbox}[boxrule=0.5pt,left=1.05pt,right=1.05pt]
    \textbf{Objective:} Given instances $\mathcal{X}$, annotators $\mathcal{A}$, and noisy class labels $\mathcal{Z}$, we aim to train a classification model \mbox{$\boldsymbol{y}_{\boldsymbol{\theta}^\star}: \Omega_X \rightarrow \Omega_Y$} with parameters $\boldsymbol{\theta}^\star \in \Theta$, which maximizes the accuracy:
    \begin{equation}
        \label{eq:objective}
        \boldsymbol{\theta}^\star \defas \argmax_{\boldsymbol{\theta} \in \Theta} \left(\mathrm{E}_{\boldsymbol{x}, \boldsymbol{y}}[\boldsymbol{y}^\mathrm{T} \boldsymbol{y}_{\boldsymbol{\theta}}(\boldsymbol{x})]\right).
    \end{equation}
\end{tcolorbox}

\section{Related Work}
\label{sec:related-work}
Learning from noisy class labels is a highly relevant research area~\cite{algan2021image,frenay2013classification,song2022learning}. Here, we focus on one- and two-stage approaches~\cite{li2022beyond} in the multi-annotator classification setup (cf.~Section~\ref{sec:problem-setup}) and robust regularization approaches~\cite{song2022learning}, to which \texttt{mixup} belongs.

\subsection{Multi-annotator Classification}
\textbf{Two-stage} multi-annotator classification approaches approximate true class labels by aggregating multiple noisy class labels per instance. Subsequently, the aggregated class labels and their associated instances serve as the training dataset for the downstream task. The simplest aggregation approach is majority voting, which outputs the class label with the most annotator votes per instance. By doing so, majority voting naively assumes all annotators have the same accuracy~\cite{chen2022label,jiang2021learning}. More advanced approaches~\cite{chen2022label,dawid1979maximum,jiang2021learning,tian2015max} overcome this issue by estimating each annotator's performance when aggregating class labels. For example, the Dawid-Skene algorithm~\cite{dawid1979maximum} estimates a confusion matrix per annotator. 
However, such approaches typically expect multiple class labels for each instance~\cite{khetan2018learning}.

\textbf{One-stage} multi-annotator classification approaches do not need multiple class labels per instance because they train the classification model without any detached stage for aggregating class labels. A common training principle is to leverage the expectation-maximization (EM) algorithm that iteratively updates the classification model's parameters and annotators' performance estimates (M-step) to accurately estimate the latent true class labels (E-step)~\cite{khetan2018learning,raykar2010learning,yang2018leveraging}. Such EM algorithms come at the price of high computational complexity and the need to decide when to switch between E- and M-steps~\cite{rodrigues2018deep}. Therefore, several approaches have been proposed to overcome these issues when training NNs. A common approach is to extend an NN-based classification model by a noise adaptation layer~\cite{rodrigues2018deep,wei2022deep}, whose parameters encode annotators' performances on top of the classification layer. Alternatively, a separate so-called annotator (performance) model is jointly trained with the classification model~\cite{cao2023learning,chu2021learning,herde2023multi,tanno2019learning,ibrahim2023deep}. In this case, both models' outputs are combined when optimizing the target loss. Alongside the training algorithm, the underlying assumptions regarding the modeling of annotator performance play a crucial role. Here, simplifications of the probabilistic graphical model in Fig.~\ref{fig:probabilistic-graphical-model} are often made, for example, by ignoring the instance dependency of the annotator performance~\cite{rodrigues2018deep,wei2022deep,tanno2019learning}. Our work follows recent one-stage approaches~\cite{cao2023learning,chu2021learning,herde2023multi}, which model annotator performance as a function of the latent true class label and the instance's features. 

\subsection{Robust Regularization}
Regularization reduces NNs' overfitting on instances with false class labels. However, common regularization approaches, such as weight decay~\cite{krogh1991simple} and dropout~\cite{srivastava2014dropout}, are often insufficient for tasks with severe class label noise~\cite{song2022learning}. Data augmentation via \texttt{mixup} is a more robust regularization approach~\cite{zhang2018mixup}. Given two randomly drawn instances $\boldsymbol{x}_n, \boldsymbol{x}_{\hat{n}} \in \mathcal{X}$ with their class labels $\boldsymbol{y}_n, \boldsymbol{y}_{\hat{n}} \in \mathcal{Y}$, a new instance-label pair for training is generated via convex combination:
\begin{align}
        \boldsymbol{\widetilde{x}} \defas \lambda \boldsymbol{x}_n + (1-\lambda) \boldsymbol{x}_{\hat{n}},
        \quad\boldsymbol{\widetilde{y}} \defas \lambda \boldsymbol{y}_n + (1-\lambda) \boldsymbol{y}_{\hat{n}},
\end{align}
where the scalar $\lambda \in [0, 1]$ is sampled from a symmetric beta distribution $\mathrm{Beta}(\alpha, \alpha)$ with the concentration parameter $\alpha \in \mathbb{R}_{>0}$. This way, \texttt{mixup} expands the training dataset by utilizing the idea that interpolating feature vectors linearly should result in linear interpolations of their targets, requiring minimal implementation and computational overhead. Meanwhile, various extensions of \texttt{mixup} have been proposed, including an extension mixing hidden states of NNs~\cite{verma2019manifold} and an extension specifically tailored for image~\cite{yun2019cutmix} or text data~\cite{sun2020mixup}. 
To the best of our knowledge, the only \texttt{mixup} extension~\cite{zhang2022identifying} for multiple annotators is proposed for opinion expression identification tasks~\cite{breck2007identifying}. Its idea is to make predictions by combining learned annotator embeddings for the same instance to simulate the annotation of an expert. Beyond the task type, our approach differs substantially from this extension by mixing class labels across different instances and annotators while explicitly modeling each annotator's performance.

\section{The Annot-Mix Approach}
\label{sec:approach}
\begin{figure}
    \begin{tikzpicture}[trapezium stretches=true]
    \tikzset{on grid=true}
        \node[] (0, 0) (image) {\includegraphics[height=1.1cm]{figures/Leopard-Jaguar.pdf}};
        \node[below=2.0 of image] (annotator){\includegraphics[height=1cm]{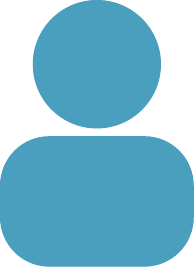}};
        \node[trapezium, trapezium left angle=80, trapezium right angle=80, right=2.0 of image, draw=black, shape border rotate=270, minimum height=1.6cm, align=center, color=pink_ground_truth] (classifier) {{\scriptsize Classification} \\ {\scriptsize Model} \\ $\boldsymbol{\theta}$};
        \node[trapezium, trapezium left angle=80, trapezium right angle=80, right=2.0 of annotator, draw=black, shape border rotate=270, align=center, minimum height=1.6cm, color=pink_ground_truth] (annotation_model) {\scriptsize Annotator \\ {\scriptsize Model} \\ $\boldsymbol{\pi}$};
        \tikzset{on grid=false}
        \node[draw=black, rounded corners=0.1cm, right= 0.4 of annotation_model, color=green_annotations] (matrix) {$\mathbf{P}_{\boldsymbol{\pi}}(\boldsymbol{h}_{\boldsymbol{\theta}}(\widetilde{\boldsymbol{x}}), \widetilde{\boldsymbol{a}})$};
        \node[rectangle, draw=black, rounded corners=0.1cm, right=0.4 of classifier, color=green_annotations] (gt_vector) {$\boldsymbol{p}^\mathrm{T}_{\boldsymbol{\theta}}(\widetilde{\boldsymbol{x}})$};
        \node[trapezium, trapezium left angle=80, trapezium right angle=80, right=0.2 of gt_vector, yshift=-0.7cm, draw=black, shape border rotate=270, align=center, color=gray] (multiply) {{\scriptsize Matrix} \\ {\scriptsize Product}};
        \node[draw=black, rounded corners=0.1cm, right=0.4 of multiply, color=green_annotations] (al_vector) {$\boldsymbol{p}^\mathrm{T}_{\boldsymbol{\theta}, \boldsymbol{\pi}}(\widetilde{\boldsymbol{x}}, \widetilde{\boldsymbol{a}})$};
        \draw[draw=gray, -latex] (image.east) -- (classifier.west);
        \draw[draw=gray, -latex] (annotator.east) -- (annotation_model.west);
        \draw[draw=gray, -latex] (classifier.east) -- (gt_vector.west);
        \draw[draw=gray, -latex] (annotation_model.east) -- (matrix.west);
        \draw[draw=gray, -latex] let  
      \p1 = (gt_vector.south),
      \p2 = (multiply.160)
    in
        (\x1, \y1) -- (\x1, \y2) -- (\x2, \y2);
        \draw[draw=gray, -latex] let  
      \p1 = (gt_vector.south),
      \p2 = (multiply.200),
      \p3 = (matrix.north),
    in
        (\x1, \y3) -- (\x1, \y2) -- (\x2, \y2);
        \draw[draw=gray, -latex] (multiply.east) -- (al_vector);

        \node[above = 0 of image]{{\scriptsize Instance} $\widetilde{\boldsymbol{x}}$};

        \node[above = 0 of annotator]{{\scriptsize Annotator} };
        \node[below = - 0.575cm of annotator]{$\widetilde{\boldsymbol{a}}$};
        \node[above = 0 of gt_vector]{{\scriptsize}};
        \draw[draw=gray, -latex] (classifier.south) -- (annotation_model.north) node[midway,right]{\textcolor{pink_ground_truth}{$\boldsymbol{h}_{\boldsymbol{\theta}}(\widetilde{\boldsymbol{x}})$}};
        \node[above = 0 of gt_vector, align=center, color=green_annotations]{\scriptsize Probabilities for \\ \scriptsize True Labels};
        \node[below = 0 of matrix, align=center, color=green_annotations]{\scriptsize Confusion Matrix};
        \node[above = 0 of al_vector, align=center, color=green_annotations]{\scriptsize Probabilities for \\ \scriptsize Noisy Labels};
        \node[] at (0, -3.05) {\scriptsize Mixed Inputs};
        \node[color=pink_ground_truth] at (1.96, -3.05) {\scriptsize Neural Networks};
        \node[color=green_annotations] at (5.25, -3.05) {\scriptsize Outputs};
    \end{tikzpicture}
    \caption{Architecture and forward propagation of \texttt{annot-mix}: The inputs, obtained after using our \texttt{mixup} extension, are propagated through the classification and annotator model, whose outputs are combined to obtain the probabilities of the observed noisy class labels.}
    \label{fig:architecture}
\end{figure}

This section presents our one-stage multi-annotator classification approach \texttt{annot-mix}, which we train through marginal likelihood maximization while leveraging our novel \texttt{mixup} extension for robust regularization.
Figure~\ref{fig:architecture} overviews our approach's architecture.

\subsection{Marginal Likelihood Maximization}
Assuming the probabilistic graphical model of Fig.~\ref{fig:probabilistic-graphical-model}, the joint distribution of the true class label $\boldsymbol{y}$ and noisy class label $\boldsymbol{z}$ given the instance $\boldsymbol{x}_n$ and the annotator $\boldsymbol{a}_m$ factors into a product of two categorical distributions:
\begin{align}
    \Pr(\boldsymbol{y}, \boldsymbol{z} \cond \boldsymbol{x}_n, \boldsymbol{a}_m) =  \Pr(\boldsymbol{y} \cond \boldsymbol{x}_n) \cdot \Pr(\boldsymbol{z} \cond \boldsymbol{x}_n, \boldsymbol{a}_m, \boldsymbol{y}).
\end{align}
For predicting instances' class labels, we aim to estimate an instance's class-membership probability distribution $\Pr(\boldsymbol{y} \cond
\boldsymbol{x}_n)$. Therefore, we employ a classification model in the form of an NN with parameters $\boldsymbol{\theta} \in \Theta$, defined through the function
\begin{align}
    \boldsymbol{p}_{\boldsymbol{\theta}}: \Omega_X \rightarrow \Delta_C \defas \{\boldsymbol{p} \in [0, 1]^C \cond ||\boldsymbol{p}||_1 = 1\},
\end{align}
where $|| \cdot ||_1$ denotes the 1-norm and $\boldsymbol{p}_{\boldsymbol{\theta}}(\boldsymbol{x}_n)$ are the estimated probabilities for the true class label of instance~$\boldsymbol{x}_n$.
Accordingly, the estimated Bayes optimal prediction for our objective in Eq.~\eqref{eq:objective} is given by the class with the maximum probability estimate:
\begin{align}
    \boldsymbol{y}_{\boldsymbol{\theta}}(\boldsymbol{x}_n) \defas \argmax_{\boldsymbol{e}_c \in \Omega_Y} \left(\boldsymbol{e}^\mathrm{T}_c\boldsymbol{p}_{\boldsymbol{\theta}}(\boldsymbol{x}_n)\right).
\end{align}
For estimating annotators' performances, we aim to approximate the probability distribution $\Pr(\boldsymbol{z} \cond \boldsymbol{x}_n, \boldsymbol{a}_m, \boldsymbol{y})$ for each possible class label $\boldsymbol{y} \in \Omega_Y$. Therefore, we employ an annotator model in the form of an NN with parameters $\boldsymbol{\pi} \in \Pi$, defined through the function
\begin{align}
    \mathbf{P}_{\boldsymbol{\pi}}: \Omega_H \times \Omega_A \rightarrow \{(\boldsymbol{p}_1, \dots, \boldsymbol{p}_C)^\mathrm{T} \cond \boldsymbol{p}_1, \dots, \boldsymbol{p}_ C \in \Delta_C\},
\end{align}
where $\mathbf{P}_{\boldsymbol{\pi}}(\boldsymbol{h}_{\boldsymbol{\theta}}(\boldsymbol{x}_n), \boldsymbol{a}_m)$ is the estimated confusion matrix of annotator~$\boldsymbol{a}_m$ for instance~$\boldsymbol{x}_n$, represented through $\boldsymbol{h}_{\boldsymbol{\theta}}(\boldsymbol{x}_n) \in \Omega_H$ as output of the classification model's penultimate layer.

Since the true class labels $\mathcal{Y}$ in the complete likelihood function $\Pr(\mathcal{Y}, \mathcal{Z} \cond \mathcal{X}, \mathcal{A}; \boldsymbol{\theta}, \boldsymbol{\pi})$ are latent, we optimize both models' parameters $(\boldsymbol{\theta}, \boldsymbol{\pi})$ by maximizing the marginal likelihood~\cite{herde2023multi} of the observed noisy class labels $\mathcal{Z}^\prime \defas \{\boldsymbol{z}_{nm} \in \mathcal{Z} \cond \boldsymbol{z}_{nm} \neq \boldsymbol{0}\}$:
\begin{align}
    \Pr(\mathcal{Z}^\prime \cond \mathcal{X}, \mathcal{A}; \boldsymbol{\theta}, \boldsymbol{\pi}) &= \prod_{\boldsymbol{x}_n \in \mathcal{X}} \prod_{\boldsymbol{a}_m \in \mathcal{A}_n} \Pr(\boldsymbol{z}_{nm} \cond \boldsymbol{x}_n, \boldsymbol{a}_m; \boldsymbol{\theta}, \boldsymbol{\pi}) \\
    \label{eq:marginalization}
    &\hspace*{-2cm}= \prod_{\boldsymbol{x}_n \in \mathcal{X}} \prod_{\boldsymbol{a}_m \in \mathcal{A}_n} 
    \left(
    \sum_{c=1}^C
    \begin{aligned}
         &\hspace*{0.65cm}\Pr(\boldsymbol{y}=\boldsymbol{e}_c \cond \boldsymbol{x}_n; \boldsymbol{\theta}) \, \cdot \\ &\Pr(\boldsymbol{z}_{nm} \cond \boldsymbol{x}_n, \boldsymbol{a}_m, \boldsymbol{y}=\boldsymbol{e}_c; \boldsymbol{\pi})
    \end{aligned}
    \right)
    \\
    \label{eq:marginal-likelihood}
    &\hspace*{-2cm}= \prod_{\boldsymbol{x}_n \in \mathcal{X}} \prod_{\boldsymbol{a}_m \in \mathcal{A}_n} \underbrace{\boldsymbol{p}^\mathrm{T}_{\boldsymbol{\theta}}(\boldsymbol{x}_n)\mathbf{P}_{\boldsymbol{\pi}}(\boldsymbol{h}_{\boldsymbol{\theta}}(\boldsymbol{x}_n), \boldsymbol{a}_m)}_{ \boldsymbol{p}^\mathrm{T}_{\boldsymbol{\theta}, \boldsymbol{\pi}}(\boldsymbol{x}_n, \boldsymbol{a}_m)\defas} \boldsymbol{z}_{nm},
\end{align}
where $\mathcal{A}_n \defas \{\boldsymbol{a}_m \in \mathcal{A} \cond \boldsymbol{z}_{nm} \neq \boldsymbol{0}\}$ comprises the annotators who provided a class label for instance $\boldsymbol{x}_n$. The marginalization (summation) of the latent true class label is shown in Eq.~\eqref{eq:marginalization}. The function $\boldsymbol{p}_{\boldsymbol{\theta}, \boldsymbol{\pi}}: \Omega_X \times \Omega_A \rightarrow \Delta_C$ in Eq.~\eqref{eq:marginal-likelihood} outputs probabilities estimating which class label an annotator will assign to an instance. Thus, the function $\boldsymbol{z}_{\boldsymbol{\theta}, \boldsymbol{\pi}}: \Omega_X \times \Omega_A \rightarrow \Omega_Y$ outputting the class label with the highest estimated probability is given by:
\begin{equation}
    \label{eq:annot-prediction}
    \boldsymbol{z}_{\boldsymbol{\theta}, \boldsymbol{\pi}}(\boldsymbol{x}_n, \boldsymbol{a}_m) \defas \argmax_{\boldsymbol{e}_c \in \Omega_Y} \left(\boldsymbol{e}_c^\mathrm{T}\boldsymbol{p}_{\boldsymbol{\theta}, \boldsymbol{\pi}}(\boldsymbol{x}_n, \boldsymbol{a}_m)\right).
\end{equation}
Converting the marginal likelihood function in Eq.~\eqref{eq:marginal-likelihood} into a negative log-likelihood function yields the cross-entropy loss function:
\begin{align}
    \label{eq:loss}
    L_{\mathcal{X}, \mathcal{A}, \mathcal{Z}^\prime}(\boldsymbol{\theta}, \boldsymbol{\pi}) \defas - \sum_{\boldsymbol{x}_n \in \mathcal{X}} \sum_{\boldsymbol{a}_m \in \mathcal{A}_n} \frac{ \boldsymbol{z}_{nm}^\mathrm{T}\ln\left(\boldsymbol{p}_{\boldsymbol{\theta}, \boldsymbol{\pi}}(\boldsymbol{x}_n, \boldsymbol{a}_m)\right)}{|\mathcal{Z}^\prime|},
\end{align}
where $\ln(\cdot)$ is applied elementwise to the input vector.
By default, optimal predictions for $\boldsymbol{p}_{\boldsymbol{\theta}}(\boldsymbol{x}_n)$ and $\mathbf{P}_{\boldsymbol{\pi}}(\boldsymbol{h}_{\boldsymbol{\theta}}(\boldsymbol{x}_n), \boldsymbol{a}_m)$ are not identifiable~\cite{ibrahim2023deep} because there are multiple combinations to produce the same output $\boldsymbol{p}_{\boldsymbol{\theta}, \boldsymbol{\pi}}(\boldsymbol{x}_n, \boldsymbol{a}_m)$. Therefore, we resort to a common solution proposed in the literature~\cite{wei2022deep,herde2023multi} and initialize the annotator model's parameters $\boldsymbol{\pi}$ to approximately satisfy:
\begin{equation}
    \forall \boldsymbol{h} \in \Omega_H, \forall \boldsymbol{a} \in \Omega_A: \mathbf{P}_{\boldsymbol{\pi}}(\boldsymbol{h}, \boldsymbol{a}) \approx \eta \boldsymbol{I}_C + \frac{\left(1-\eta\right)}{C-1} \left(\boldsymbol{1}_C - \boldsymbol{I}_C\right),
\end{equation}
with $\eta \in (0, 1)$ as the probability of obtaining a correct class label, $\boldsymbol{I}_C \in \mathbb{R}^{C \times C}$ as an identity matrix, and $\boldsymbol{1}_C \in \mathbb{R}^{C \times C}$ as an all-one matrix. We set $\eta \defas 0.9 > \sfrac{1}{C}$, implying that solutions with diagonally dominant confusion matrices are preferred at training start. 

\subsection{Mixup Extension}
Optimizing the loss function in Eq.~\eqref{eq:loss} corresponds to empirical risk minimization (ERM)~\cite{vapnik1995the} because the classification and annotator models are forced to fit the observed noisy class labels perfectly. Although the annotator model attempts to separate the noise in the class labels during training, overfitting is still an issue (cf. Section~\ref{sec:evaluation}). Therefore, we extend \texttt{mixup}~\cite{zhang2018mixup} for robust regularization and improved generalization of \texttt{annot-mix}. Our idea for the extension of \texttt{mixup} to multi-annotator classification tasks lies in shifting the perspective from mixing tuples of instances and class labels to mixing triples of instances, annotators, and class labels. Concretely, we propose the following extension:
\begin{tcolorbox}[boxrule=0.5pt,left=1.05pt,right=1.05pt]
    Given two triples $(\boldsymbol{x}_n, \boldsymbol{a}_m, \boldsymbol{z}_{nm}), (\boldsymbol{x}_{\hat{n}}, \boldsymbol{a}_{\hat{m}}, \boldsymbol{z}_{{\hat{n}}{\hat{m}}})$, randomly sampled from
    \begin{equation}
        \label{eq:triples}
        \mathcal{M} \defas \{(\boldsymbol{x}_n, \boldsymbol{a}_m, \boldsymbol{z}_{nm}) \cond \boldsymbol{x}_n \in \mathcal{X}, \boldsymbol{a}_m \in \mathcal{A}, \boldsymbol{z}_{nm} \in \mathcal{Z}^\prime\},
    \end{equation}
    we apply \texttt{mixup} to instances, annotators, and noisy labels via
   \begin{align}
        \label{eq:mixinstances}
        \widetilde{\boldsymbol{x}} &\defas \lambda \boldsymbol{x}_{n} + (1-\lambda)\boldsymbol{x}_{\hat{n}}, \\
        \label{eq:mixannotators}
        \widetilde{\boldsymbol{a}} &\defas \lambda \boldsymbol{a}_{m} + (1-\lambda)\boldsymbol{a}_{\hat{m}}, \\
        \label{eq:mixlabels}
        \widetilde{\boldsymbol{z}} &\defas \lambda \boldsymbol{z}_{nm} + (1-\lambda)\boldsymbol{z}_{\hat{n}\hat{m}},\\
        \label{eq:coefficient}
        \lambda &\sim \mathrm{Beta}(\alpha, \alpha), \alpha > 0.
    \end{align}  
\end{tcolorbox}
\begin{figure}[!ht]
\centering
\begin{tikzpicture}[x=1.8cm,y=1.0cm]
  \draw[->] (0, 0) -- (3, 0) node[right] {$\Omega_{X}$};
  \draw[->] (0, 0) -- (0, 3) node[above] {$\mathrm{conv}(\Omega_{A}$)};
  \draw (0.5,-0.1) node(feat_1)[below]{\includegraphics[height=1 cm]{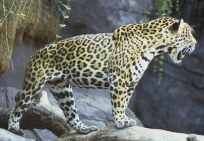}} -- +(0,0.2);
  \draw (1.5,-0.1) node(feat_2)[below]{\includegraphics[height=1 cm]{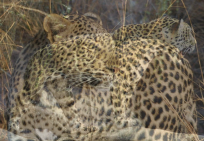}} -- +(0,0.2);
  \draw (2.5,-0.1) node(feat_3)[below]{\includegraphics[height=1 cm]{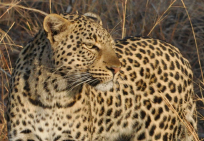}} -- +(0,0.2);

  \draw (-0.1, 0.5) node(annot_1)[left]{\includegraphics[height=.9 cm]{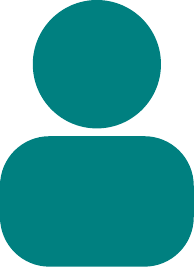}} -- +(0.2,0);
  \draw (-0.1, 1.5) node(annot_2)[left]{\includegraphics[height=.9 cm]{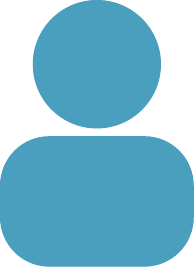}} -- +(0.2,0);
  \draw (-0.1, 2.5) node(annot_3)[left]{\includegraphics[height=.9 cm]{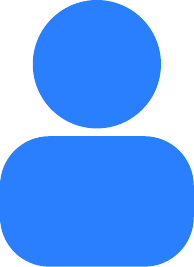}} -- +(0.2,0);

  \node[below = - 0.53cm of annot_1]{$\boldsymbol{a}_{m}$};
  \node[below = - 0.575cm of annot_2]{$\widetilde{\boldsymbol{a}}$};
  \node[below = - 0.53cm of annot_3]{$\boldsymbol{a}_{\hat{m}}$};

  \node[below = - 0.1cm of feat_1]{$\boldsymbol{x}_{n}$};
  \node[below = - 0.15cm of feat_2]{$\widetilde{\boldsymbol{x}}$};
  \node[below = - 0.1cm of feat_3]{$\boldsymbol{x}_{\hat{n}}$};

  \draw[blue,  path fading=east] (0.5, 2.5) node(a){} 
  -- (1.5, 1.5) node(b){}
  -- (2.5, 0.5) node(c){};

  \draw[red,  path fading=west] (0.5, 2.5) node(a){} 
  -- (1.5, 1.5) node(b){}
  -- (2.5, 0.5) node(c){};

  \fill[fill=blue] (a) circle[radius=1pt];
  \fill[fill=blue!50!red] (b) circle[radius=1pt];
  \fill[fill=red] (c) circle[radius=1pt];

  \node[above right = - 0.15cm of a]{$\boldsymbol{z}_{nm}$};
  \node[above right = - 0.15cm of b]{$\widetilde{\boldsymbol{z}}$};
  \node[above right = - 0.15cm of c]{$\boldsymbol{z}_{{\hat{n}{\hat{m}}}}$};

  \node[below left=0.05cm of b]{\textcolor{blue!50!red}{$\lambda=0.5$}};
\end{tikzpicture}
\caption{\texttt{mixup} extension to multi-annotator classification: We convexly combine class labels from (potentially) different annotators and instances and thus augment data in the instance and annotator feature space.}
\label{fig:mix-up-illustration}
\end{figure}
Applying the above formulation in the convex hull of the annotator feature space, $\operatorname{conv}(\Omega_A)$, allows us to handle varying numbers of noisy class labels per instance while preserving information about each label's annotator. Moreover, we can natively manage even datasets with only one class label for each instance. Figure~\ref{fig:mix-up-illustration} illustrates our \texttt{mixup} extension as data augmentation in the instance feature space and the convex hull of the annotator feature space. Intuitively, this has two main effects. On the one hand, we simultaneously regularize the classification and annotator models. This is because mixing class labels from different annotators across instances makes it more difficult to memorize which class label an annotator provides for an instance. On the other hand, we improve generalization by linearly interpolating not only in the instance feature space but also in the annotator feature space. We demonstrate both effects in our ablation and hyperparameter study (cf. Section~\ref{sec:evaluation}) with an analysis of the $\alpha$ hyperparameter controlling the degree of regularization. For example, defining $\alpha \rightarrow 0$ recovers the ERM solution.

\subsection{Implementation}
Figure~\ref{fig:python-code} summarizes the training with \texttt{annot-mix} as a Python code snippet. The design of the classification model's architecture depends on the underlying data modality and task. For example, one may employ a residual network (ResNet)~\cite{he2016deep} for image data. In contrast, the annotator model must process two vectors as inputs. For this purpose, we use a simple multi-layer perceptron (MLP) with input concatenation. Our repository provides more implementation details.\footnoteref{fn:repo}
\begin{figure}[!h]
\centering
\footnotesize
\begin{minted}[mathescape, linenos=false, texcomments]{python}
# Build data loaders for the set $\mathcal{M}$ (cf. Eq. \eqref{eq:triples}).
loaders = zip(loader1, loader2)
# Iterate over the randomly shuffled data from 
# both data loaders in batches.
for (x1, a1, z1), (x2, a2, z2) in loaders:
    # Sample mixing coefficient (cf. Eq. \eqref{eq:coefficient}).
    lmbda = np.random.beta(alpha, alpha)
    # Perform \texttt{mixup} (cf. Eqs. \eqref{eq:mixinstances}, \eqref{eq:mixannotators}, \eqref{eq:mixlabels}).
    x = lmbda * x1 + (1-lmbda) * x2
    a = lmbda * a1 + (1-lmbda) * a2
    z = lmbda * z1 + (1-lmbda) * z2
    # Jointly optimize the classification model's  
    # parameters $\boldsymbol{\theta}$ and the annotator model's 
    # parameters $\boldsymbol{\pi}$ (cf. Eq. \eqref{eq:loss}).
    optimizer.zero_grad()
    loss = -(z * net(x, a).log()).sum() / len(z)
    loss.backward()
    optimizer.step()    
\end{minted}
\caption{Illustrative Python code snippet for one epoch of training with \texttt{annot-mix}: The NN architectures are implemented through a PyTorch (tested with \texttt{v2.1.0}) module \texttt{net}, which takes the tensors of mixed instances and annotators as input to minimize the cross-entropy loss regarding the tensors of mixed noisy class labels.}
\label{fig:python-code}
\end{figure}

\section{Empirical Evaluation}

\label{sec:evaluation}
Our empirical evaluation comprises three parts. First, we explain the basic setup of our experiments. Second, we present the results of an ablation and hyperparameter study. Third, we compare the performance of \texttt{annot-mix} to mostly state-of-the-art multi-annotator classification approaches. Code and further details on how to reproduce all experiments are available in our repository.\footnote{\label{fn:repo}Our GitHub repository is accessible via \github.}

\subsection{Experimental Setup}
We design experiments according to the problem setup of Section~\ref{sec:problem-setup}. Table~\ref{tab:datasets} overviews our setup, which we detail in the following.

\textbf{Datasets:} We select our datasets to cover a wide range of real-world settings for obtaining meaningful assessments of the approaches' robustness and performances. Concretely, experiments are performed on eleven real-world datasets across three data modalities: image, tabular, and text. The number of classes ranges from $C=4$ to $C=1{,}000$. Five datasets contain noisy class labels from humans, while we simulate annotators providing noisy class labels for the remaining six datasets. The number of annotators ranges from $M=20$ to $M=733$. As labels are costly, the average number of provided class labels per instance (approximately ranging from one to four) is considerably lower than the number of annotators. The fraction of false class labels, i.e., noise levels, ranges from low noise (ca. \SI{20}{\%}) to moderate noise (ca. \SI{40}{\%}) to high noise (ca. \SI{80}{\%}) levels.

\textbf{Annotator simulation}: 
Due to the costly annotation process, the number of publicly available datasets annotated by multiple error-prone humans is limited. Therefore, we include datasets with simulated annotators. Ideally, the simulated noisy class labels are close to human noisy class labels. To do so, we follow related work~\cite{gu2023instance} and train an individual NN for each annotator. These NNs differ in their training hyperparameters and in the training data they use. Specifically, we train $M=20$ NNs with different parameter initializations, numbers of training epochs, learning rates, and ratios of randomly sampled training instance-label pairs per class. Then, each NN's predictions serve as the noisy class labels. For example, in a binary classification problem, one NN may train with many instance-label pairs from both classes but only for a few epochs. In contrast, another NN may train for more epochs with fewer instance-label pairs of the positive class. This way, we mimic annotators with different expertise regarding certain classes and regions in the instance feature space. Obviously, we have access to the noisy class label of each simulated annotator for each instance. Yet, we set the average number of class labels per instance to a much lower number (three or one) to account for the limited annotation budget in real applications. Moreover, it is also common for some annotators to provide many labels while other annotators provide very few labels. We address this issue by adopting an existing method~\cite{zhang2024coupled}, where each annotator is assigned an individual probability for annotating an instance. This probability is proportional to the sampled value from a Beta distribution, which we parameterize as $\mathrm{Beta}(1.0, 3.0)$.

\textbf{Multi-annotator classification approaches:} For benchmarking the performance of \texttt{annot-mix}, we compare it to eight one-stage multi-annotator classification approaches, which are \texttt{crowd-layer}~\cite{rodrigues2018deep}, trace-regularized estimation of annotator confusion (\texttt{trace-reg})~\cite{tanno2019learning}, common noise adaptation layers (\texttt{conal})~\cite{chu2021learning}, learning from multiple annotators as a union (\texttt{union-net})~\cite{wei2022deep}, multi-annotator deep learning (\texttt{madl})~\cite{herde2023multi}, geometry-regularized crowdsourcing networks (\texttt{geo-reg-w}, \texttt{geo-reg-f})~\cite{ibrahim2023deep}, and learning from crowds with annotation reliability (\texttt{crowd-ar})~\cite{cao2023learning}. These one-stage approaches mainly differ regarding their training algorithms and estimation of annotators' performances. While prioritizing one-stage approaches for their reported performance gains~\cite{ibrahim2023deep}, we still include basic two-stage approaches to better contextualize the results. Specifically, we employ majority voting (\texttt{mv-base}) as a lower baseline and combine vanilla \texttt{mixup}~\cite{zhang2018mixup} with majority voting (\texttt{mv-mixup}) and the Dawid-Skene algorithm~\cite{dawid1979maximum} (\texttt{ds-mixup}) as more advanced two-stage approaches. Further, we show the results for training with the true class labels (\texttt{true-base}) as an upper baseline.

\textbf{Evaluation scores:} According to our objective in Eq.~\eqref{eq:objective}, we assess a classification model with parameters $\boldsymbol{\theta}$ through its empirical classification accuracy on a separate test set $\mathcal{T} \subset \Omega_X \times \Omega_Y$:
\begin{equation}
    \label{eq:clf-acc}
    \texttt{clf-acc}_{\mathcal{T}}(\boldsymbol{\theta}) \defas \frac{1}{|\mathcal{T}|}\sum_{(\boldsymbol{x}_\mathit{t}, \boldsymbol{y}_\mathit{t}) \in \mathcal{T}} \boldsymbol{y}_\mathit{t}^\mathrm{T}\boldsymbol{y}_{\boldsymbol{\theta}}(\boldsymbol{x}_\mathit{t}),
\end{equation}
where the instance-label pairs in $\mathcal{T}$ are independently sampled from the joint distribution $\Pr(\boldsymbol{x}, \boldsymbol{y})$. Going beyond the standard classification setting, we additionally assess the annotator model with parameters $\boldsymbol{\pi}$. For this purpose, we adopt the idea of evaluating how well the model can predict whether an annotator provides a wrong or correct class label for a certain instance~\cite{herde2023multi}. In case of $\texttt{annot-mix}$, we define a function $p_{\boldsymbol{\theta}, \boldsymbol{\pi}}: \Omega_X \times \Omega_A \rightarrow [0, 1]$, which outputs the estimated probability of obtaining a correct class label from an annotator~$\boldsymbol{a}_m$ for a given instance~$\boldsymbol{x}_n$:
\begin{equation}
    \label{eq:correctness-probability}
    p_{\boldsymbol{\theta}, \boldsymbol{\pi}}(\boldsymbol{x}_n, \boldsymbol{a}_m) \defas \boldsymbol{p}^\mathrm{T}_{\boldsymbol{\theta}}(\boldsymbol{x}_n)\mathrm{diag}\left(\mathbf{P}_{\boldsymbol{\pi}}(\boldsymbol{h}_{\boldsymbol{\theta}}(\boldsymbol{x}_n), \boldsymbol{a}_m)\right),
\end{equation}
where $\mathrm{diag}\left(\mathbf{P}_{\boldsymbol{\pi}}(\boldsymbol{h}_{\boldsymbol{\theta}}(\boldsymbol{x}_n), \boldsymbol{a}_m)\right) \in [0, 1]^C$ denotes the diagonal of the confusion matrix as a column vector.
Other one-stage multi-annotator classification approaches similarly provide performance estimates of annotators. Since this estimation task can be interpreted as a binary classification task, we compute the area under the receiver operating characteristic~\cite{huang2005using} (\texttt{perf-auroc}) to assess how well the different approaches can predict annotators' performances. Another evaluation score of interest is the accuracy in predicting the noisy class labels provided by the annotators for the training data:
\begin{equation}
    \label{eq:annot-acc}
    \texttt{annot-acc}_{\mathcal{M}}(\boldsymbol{\theta}, \boldsymbol{\pi}) \defas \frac{1}{|\mathcal{M}|}\hspace{0.8cm}\sum_{\mathclap{(\boldsymbol{x}_\mathit{n}, \boldsymbol{a}_\mathit{m}, \boldsymbol{z}_\mathit{nm}) \in \mathcal{M}}}\hspace{0.6cm} \boldsymbol{z}_{nm}^\mathrm{T}\boldsymbol{z}_{\boldsymbol{\theta}, \boldsymbol{\pi}}(\boldsymbol{x}_\mathit{n}, \boldsymbol{a}_\mathit{m}),
\end{equation}
which allows us to identify overfitting by comparing it to \texttt{clf-acc}.

\begin{table*}[!t]
        \centering
        \scriptsize
        \setlength{\tabcolsep}{4.1pt}
        \def\arraystretch{1.22}
        \caption{Experimental setup: Column headings indicate the names of the eleven datasets and rows refer to data properties. We denote numbers by prefixing them with the $\#$ symbol and indicate averages by $\overline{\#}$. The statistics of \texttt{cifar10h} and \texttt{cifar10n} refer to annotation subsets of the original datasets.}
        \label{tab:datasets}
        \begin{tabular}{|l|ccccc|cccccc|}
            \toprule
            \multicolumn{1}{|c|}{\multirow{2}{*}{\textBF{Setup}}} & \texttt{mgc} & \texttt{labelme} & \texttt{cifar10h} & \texttt{cifar10n} & \texttt{cifar100n} & \texttt{letter} & \texttt{flowers102} & \texttt{trec6} & \texttt{aloi} & \texttt{dtd} & \texttt{agnews}  \\
            \cmidrule(lr){2-2} \cmidrule(lr){3-3} \cmidrule(lr){4-4} \cmidrule(lr){5-6} \cmidrule(lr){7-7} \cmidrule(lr){8-8} \cmidrule(lr){9-9} \cmidrule(lr){10-10} \cmidrule(lr){11-11} \cmidrule(lr){12-12}
            & \cite{rodrigues2013learning} & \cite{rodrigues2017learning} & \cite{peterson2019human}  & \multicolumn{2}{c|}{\cite{wei2021learning}} & \cite{frey1991letter} & \cite{nilsback2008automated} & \cite{li2002learning} & \cite{geusebroek2005amsterdam} & \cite{cimpoi2014describing} & \cite{zhang2015character} \\
            \hline
            \multicolumn{12}{|c|}{\cellcolor{datasetcolor!10} General}\\
            \hline
            data modality & tabular & image & image & image & image & tabular & image  & text & tabular & image & text \\
            $\#$ training instances & 700 & 1{,}000 & 8{,}621 & 50{,}000 & 50{,}000 & 15{,}500 & 1{,}020 & 4,952 & 84{,}400 & 1{,}880 & 118{,}000 \\
            $\#$ validation instances & 100 & 500 & 500 & 500 & 500 & 500 & 1{,}020 & 500 & 2{,}000 & 1{,}880 & 2{,}000 \\
            $\#$ test instances & 200 & 1{,}188 & 49{,}500 & 9{,}500 & 9{,}500 & 4{,}000 & 6{,}149 & 500 & 21{,}600 & 1{,}880 & 7{,}600  \\
            $\#$ classes & 10 & 8 & 10 & 10 & 100 & 26 & 102 & 6  & 1{,}000 & 47 & 4 \\
            \hline
            \multicolumn{12}{|c|}{\cellcolor{datasetcolor!10} Annotations} \\
            \hline
            human annotators & \ding{51} & \ding{51} & \ding{51} & \ding{51} & \ding{51} & \ding{55} & \ding{55} & \ding{55}  & \ding{55} & \ding{55} & \ding{55} \\
            $\#$ annotators & 44 & 59 & 100 & 733 & 519 & 20 & 20 & 20  & 20 & 20 & 20 \\
            $\overline{\#}$ class labels per instance & 4.2 & 2.5 & 2.3 & 1.0 & 1.0 & 3.0 & 3.0 & 3.0 & 1.0 & 1.0 & 1.0 \\
            $\overline{\#}$ class labels per annotator & 70 & 43 & 200 & 68 & 96 & 2{,}325 & 152 & 743 & 4{,}220 & 94 & 5{,}900 \\
            $\%$ false class labels & 44.0 & 26.0 & 22.5 & 40.2 & 40.2 & 51.9 & 67.5 & 36.9 & 43.4 & 76.8 & 56.8\\
            \hline
            \multicolumn{12}{|c|}{\cellcolor{datasetcolor!10} Training} \\
            \hline
            architecture & MLP & DINOv2 & ResNet18 & ResNet18 & ResNet18 & MLP & DINOv2 & BERT  & MLP & DINOv2 & BERT \\
            pretrained & \ding{55} & \ding{51} & \ding{55} & \ding{55} & \ding{55} & \ding{55} & \ding{51} & \ding{51} & \ding{55} & \ding{51} & \ding{51} \\
            \# epochs &  50 & 50 & 100 & 100 & 100 & 50 & 50  & 50  & 50 & 50 & 50 \\
            optimizer & RAdam & RAdam & RAdam & RAdam & RAdam & RAdam & RAdam  & RAdam  & RAdam & RAdam & RAdam \\
            batch size & 64 & 64 & 128 & 128 & 128 & 64 & 64  & 64 & 64 & 64 & 64\\
            learning rate & 1e-2 & 1e-2 & 1e-3 & 1e-3 & 1e-3 & 1e-2 & 1e-2 & 1e-2 & 1e-2  & 1e-2 & 1e-2 \\
            weight decay  & 0 & 1e-4 & 1e-4 & 1e-4 & 1e-4 & 0 & 1e-4 & 1e-4 & 0 & 1e-4 & 1e-4 \\
            \bottomrule
        \end{tabular}
\end{table*}

\textbf{Architectures}: We specify architectures to meet the requirements of the respective datasets. For the three tabular datasets \texttt{mgc}, \texttt{letter}, and \texttt{aloi}, we train a simple MLP with two hidden layers. For the datasets \texttt{cifar10h}, \texttt{cifar10n}, and \texttt{cifar100n}, which consist of $32\times32$ images, we employ a ResNet18~\cite{he2016deep}. The other three image datasets \texttt{labelme}, \texttt{flowers102}, and \texttt{dtd} contain higher-resolution images, so we use DINOv2~\cite{oquab2023dinov2} as a pre-trained vision transformer (ViT). More concretely, we freeze the feature extraction layers of the ViT-S/14 and train the classification head implemented through an MLP with one hidden layer. Typical image data augmentations are performed for the six image datasets. An analogous procedure is applied to the text datasets \texttt{trec6} and \texttt{agnews}, with the difference that we use bidirectional encoder representations from transformers (BERT)~\cite{devlin2019bert} as a pre-trained architecture. 

\textbf{Training:} For all datasets, we employ RAdam~\cite{liu2019variance} as the optimizer and cosine annealing~\cite{loshchilov2017sgdr} as the learning rate scheduler. Training hyperparameters (cf. Table~\ref{tab:datasets}), such as the initial learning rate, the batch size, the number of training epochs, and weight decay, are empirically specified to ensure proper learning and convergence of the \texttt{true-base}. We set further hyperparameters specific to a multi-annotator classification approach according to the recommendations of the respective authors. This way, we ensure meaningful and fair comparisons. Moreover, training, validation, and test sets are given for each dataset. If no validation set is provided by the respective data creators, we define a small validation set with true class labels. In this case, the validation size is set either to $100$, $500$, or $2{,}000$, depending on the number $N$ of training instances. Following related works~\cite{ruhling2021end,zhang2024coupled}, we use such a validation set to select the model parameters with the highest validation accuracy throughout the training epochs. However, acquiring a validation set with true class labels may be costly in settings with noisy class labels~\cite{yuan2024early}. Thus, we also report the results for the models obtained after the last training epoch. Each experiment is repeated ten times with different parameter initializations. Accordingly, all results refer to means and standard deviations over these ten repetitions. 

\subsection{Ablation and Hyperparameter Study}
This study analyzes the regularization and generalization effect of our \texttt{mixup} extension as a part of \texttt{annot-mix}. Further, we study the gain of mixing annotators and class labels across different instances.

\begin{figure}[!t]
    \centering
    \begin{tikzpicture}
        \node(x-axis) at (0, 0) {\rotatebox{90}{{\phantom{0}\color{color_annot_violet}\scriptsize \texttt{clf-acc} [\%]} | {\color{color_annot_green}\scriptsize \texttt{annot-acc} [\%]}}};
        \node[right = 3.95cm of x-axis] {\rotatebox{90}{{\phantom{0}\color{color_annot_violet}\scriptsize \texttt{clf-acc} [\%]} | {\color{color_annot_green}\scriptsize \texttt{annot-acc} [\%]}}};
        \node(learning-curves)[above right = -4.45cm and -0.1cm of x-axis] {\includegraphics[width=0.45\textwidth]{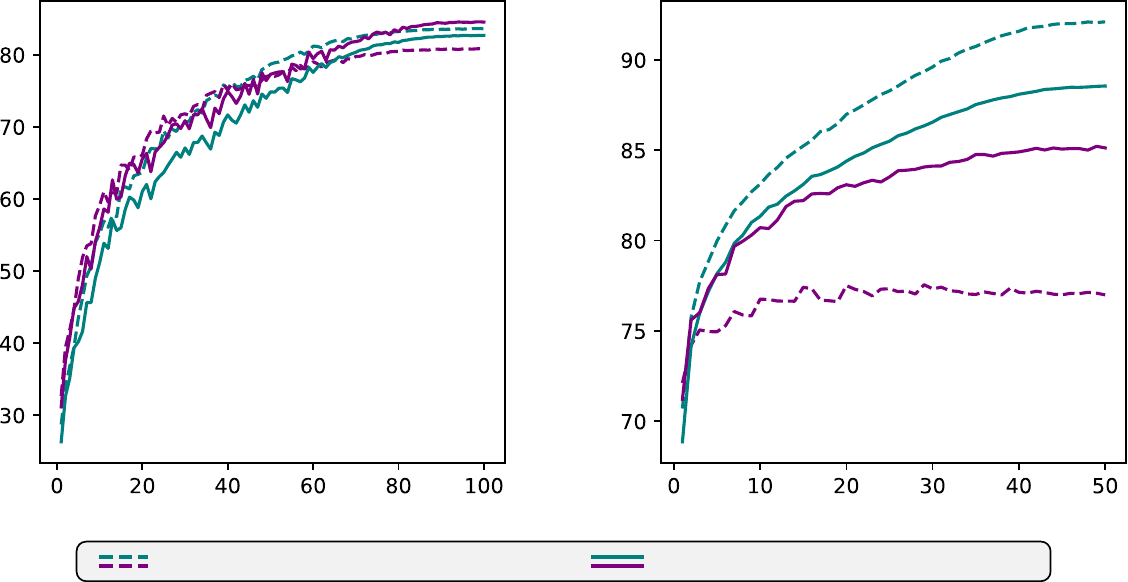}};
        \node[below left = -0.82cm and -2.675cm of learning-curves] {\scriptsize $\#$ Epochs};
        \node[below right = -0.82cm and -2.475cm of learning-curves] {\scriptsize $\#$ Epochs};
        \node(title)[above left =-0.15cm and -2.6cm of learning-curves] {\scriptsize \texttt{cifar10h}};
        \node(title)[above right =-0.15cm and -2.3cm of learning-curves] {\scriptsize \texttt{letter}};
        \node(legend-1) at (2.85, -2.22) {\parbox{2.7cm}{\scriptsize \texttt{annot-mix} w/o \texttt{mixup}}};
        \node(legend-4) at (6.35, -2.22) {\parbox{2.7cm}{\scriptsize \texttt{annot-mix} w/ \texttt{mixup}}};
    \end{tikzpicture}
    \caption{Exemplary learning curves of \texttt{annot-mix} with (w/) and without (w/o) \texttt{mixup} for the datasets \texttt{cifar10h} and \texttt{letter}.}
    \label{fig:learning-curves}
\end{figure}

\textbf{Regularization effect:} Figure~\ref{fig:learning-curves} exemplarily depicts the learning curves of \texttt{annot-mix} with ($\alpha=1$, solid line) and without ($\alpha \rightarrow 0$, dashed line) our \texttt{mixup} extension for the datasets \texttt{cifar10h} and \texttt{letter}. The colors distinguish the two evaluation scores \texttt{clf-acc} (test set) and \texttt{annot-acc} (training set). The observation that the greenish dashed learning curves surpass the greenish solid learning curves demonstrates that training \texttt{annot-mix} with our \texttt{mixup} extension diminishes the accuracy of predicting noisy class labels assigned by annotators within the training set. In other words, our \texttt{mixup} extension makes memorizing the training data more difficult. Yet, the observation that the purplish dashed learning curves fall short of the purplish solid learning curves demonstrates that our \texttt{mixup} extension boosts the test accuracy. Together, these observations verify our \texttt{mixup} extension reduces overfitting to noisy labels. 

\begin{table}[!ht]
    \setlength{\tabcolsep}{1.22pt}
    \def\arraystretch{1.22}
    \scriptsize
    \caption{
        Ablation and hyperparameter study of \texttt{annot-mix}: \best{Best} and \secondbest{second best} results for the \texttt{clf-acc} [\%] (cf.~Eq.~\eqref{eq:clf-acc}) are marked per dataset (row-wise). Numbers right to the datasets (second column) indicate false label fractions [\%].
    }
    \label{tab:ablation}
    \centering
    \begin{tabular}{|r|c|cccccc|}
        \toprule
        \multicolumn{2}{|c|}{\multirow{2}{*}{\textBF{\texttt{clf-acc} [\%]}}} & \multicolumn{1}{c}{w/o \texttt{mixup}}   & \multicolumn{4}{c}{w/ \texttt{mixup}} & idea \cite{zhang2022identifying}\\
        \cmidrule(lr){3-3} \cmidrule(lr){4-7} \cmidrule(lr){8-8}
        \multicolumn{2}{|c|}{} & $\alpha \rightarrow 0.0$ & $\alpha=0.5$ & $\alpha=1.0$ & $\alpha=2.0$ & $\alpha=4.0$ & $\alpha=1.0$ \\
        \hline
        \multicolumn{8}{|c|}{\cellcolor{datasetcolor!10} Last Epoch}\\
        \hline
        \texttt{mgc} & 44.0 & \best{$\text{74.5}_{\pm \text{1.7}}$} & \secondbest{$\text{74.2}_{\pm \text{1.7}}$} & $\text{73.8}_{\pm \text{1.1}}$ & $\text{73.3}_{\pm \text{1.5}}$ & $\text{72.7}_{\pm \text{1.5}}$ & $\text{72.4}_{\pm \text{1.5}}$ \\
        \texttt{labelme} & 26.0 & $\text{83.7}_{\pm \text{2.0}}$ & $\text{85.7}_{\pm \text{0.6}}$ & \secondbest{$\text{85.8}_{\pm \text{0.6}}$} & \best{$\text{86.0}_{\pm \text{0.4}}$} & $\text{85.4}_{\pm \text{0.5}}$ & $\text{83.6}_{\pm \text{0.5}}$\\
        \texttt{cifar10h} & 22.5 & $\text{80.8}_{\pm \text{0.3}}$ & $\text{83.8}_{\pm \text{0.2}}$ & $\text{84.6}_{\pm \text{0.2}}$ & \best{$\text{85.0}_{\pm \text{0.2}}$} & \secondbest{$\text{84.9}_{\pm \text{0.2}}$} & $\text{80.8}_{\pm \text{0.2}}$ \\
        \texttt{cifar10n} & 40.2 & $\text{72.7}_{\pm \text{0.3}}$ & $\text{79.8}_{\pm \text{0.2}}$ & $\text{82.4}_{\pm \text{0.5}}$ & \secondbest{$\text{84.8}_{\pm \text{0.2}}$} & \best{$\text{85.3}_{\pm \text{0.2}}$} & $\text{72.7}_{\pm \text{0.3}}$ \\
        \texttt{cifar100n} & 40.2 & $\text{59.4}_{\pm \text{0.4}}$ & $\text{63.4}_{\pm \text{0.4}}$ & \secondbest{$\text{64.7}_{\pm \text{0.4}}$} & \best{$\text{65.4}_{\pm \text{0.3}}$} & $\text{64.6}_{\pm \text{0.3}}$ & $\text{59.4}_{\pm \text{0.4}}$ \\
        \hline
        \texttt{letter} & 51.9 & $\text{76.6}_{\pm \text{1.8}}$ & \best{$\text{85.5}_{\pm \text{1.2}}$} & \secondbest{$\text{85.1}_{\pm \text{1.3}}$} & $\text{84.7}_{\pm \text{0.5}}$ & $\text{83.1}_{\pm \text{0.4}}$ & $\text{82.6}_{\pm \text{2.0}}$   \\
        \texttt{flowers102} & 67.5 & $\text{88.4}_{\pm \text{1.1}}$ & \best{$\text{90.6}_{\pm \text{1.2}}$} & \secondbest{$\text{90.1}_{\pm \text{0.9}}$} & $\text{89.8}_{\pm \text{1.2}}$ & $\text{89.2}_{\pm \text{1.1}}$ & $\text{89.5}_{\pm \text{1.0}}$ \\
        \texttt{trec6} & 36.9 & \best{$\text{93.3}_{\pm \text{0.7}}$} & $\text{92.1}_{\pm \text{0.4}}$ & \secondbest{$\text{92.3}_{\pm \text{0.6}}$} & $\text{91.7}_{\pm \text{0.7}}$ & $\text{91.6}_{\pm \text{0.6}}$ & $\text{92.2}_{\pm \text{0.7}}$\\
        \texttt{aloi} & 43.4 & $\text{80.7}_{\pm \text{0.1}}$ & \best{$\text{83.8}_{\pm \text{0.1}}$} & \secondbest{$\text{83.6}_{\pm \text{0.1}}$} & $\text{83.1}_{\pm \text{0.1}}$ & $\text{82.5}_{\pm \text{0.1}}$ & $\text{80.7}_{\pm \text{0.1}}$ \\
        \texttt{dtd} & 76.8 & $\text{52.9}_{\pm \text{1.1}}$ & \best{$\text{55.6}_{\pm \text{1.7}}$} & $\text{55.2}_{\pm \text{1.1}}$ & \secondbest{$\text{55.3}_{\pm \text{1.2}}$} & $\text{54.0}_{\pm \text{1.0}}$ & $\text{52.9}_{\pm \text{1.1}}$  \\
        \texttt{agnews} & 56.8 & $\text{77.9}_{\pm \text{4.6}}$ & $\text{83.1}_{\pm \text{2.3}}$ & $\text{86.2}_{\pm \text{1.5}}$ & \secondbest{$\text{86.5}_{\pm \text{1.4}}$} & \best{$\text{86.8}_{\pm \text{1.4}}$} & $\text{77.9}_{\pm \text{4.6}}$  \\
        \hline
        \multicolumn{8}{|c|}{\cellcolor{datasetcolor!10} Best Epoch}\\
        \hline
        \texttt{mgc} & 44.0 & $\text{72.8}_{\pm \text{1.8}}$ & $\text{73.0}_{\pm \text{2.1}}$ & \best{$\text{73.8}_{\pm \text{2.1}}$} & \secondbest{$\text{73.2}_{\pm \text{1.5}}$} & $\text{72.7}_{\pm \text{2.2}}$ & $\text{71.5}_{\pm \text{1.7}}$ \\
        \texttt{labelme} & 26.0 & $\text{86.1}_{\pm \text{0.7}}$ & \best{$\text{86.8}_{\pm \text{0.6}}$} & \secondbest{$\text{86.5}_{\pm \text{0.7}}$} & $\text{86.0}_{\pm \text{0.4}}$ & $\text{85.6}_{\pm \text{0.7}}$ & $\text{86.2}_{\pm \text{0.8}}$ \\
        \texttt{cifar10h} & 22.5 & $\text{80.2}_{\pm \text{0.6}}$ & $\text{83.4}_{\pm \text{0.4}}$ & $\text{84.4}_{\pm \text{0.3}}$ & \secondbest{$\text{84.5}_{\pm \text{0.6}}$} & \best{$\text{84.7}_{\pm \text{0.4}}$} & $\text{80.5}_{\pm \text{0.6}}$  \\
        \texttt{cifar10n} & 40.2 & $\text{81.0}_{\pm \text{0.6}}$ & $\text{82.7}_{\pm \text{0.5}}$ & $\text{83.2}_{\pm \text{0.3}}$ & \secondbest{$\text{84.7}_{\pm \text{0.3}}$} & \best{$\text{85.4}_{\pm \text{0.2}}$} & $\text{81.0}_{\pm \text{0.6}}$ \\
        \texttt{cifar100n} & 40.2 & $\text{58.0}_{\pm \text{1.4}}$ & $\text{62.7}_{\pm \text{0.5}}$ & \secondbest{$\text{64.1}_{\pm \text{0.6}}$} & \best{$\text{64.7}_{\pm \text{0.6}}$} & $\text{64.0}_{\pm \text{0.6}}$ & $\text{58.0}_{\pm \text{1.4}}$ \\
        \hline
        \texttt{letter} & 51.9 & $\text{77.5}_{\pm \text{1.7}}$ & \best{$\text{85.1}_{\pm \text{1.2}}$} & \secondbest{$\text{84.9}_{\pm \text{1.6}}$} & $\text{84.8}_{\pm \text{0.5}}$ & $\text{83.1}_{\pm \text{0.7}}$ & $\text{82.4}_{\pm \text{2.5}}$  \\
        \texttt{flowers102} & 67.5 & $\text{88.4}_{\pm \text{1.2}}$ & \best{$\text{90.5}_{\pm \text{1.2}}$} & \secondbest{$\text{90.2}_{\pm \text{0.9}}$} & $\text{89.8}_{\pm \text{1.2}}$ & $\text{89.1}_{\pm \text{1.1}}$ & $\text{89.5}_{\pm \text{1.1}}$ \\
        \texttt{trec6} & 36.9 &  \secondbest{$\text{91.8}_{\pm \text{1.2}}$} & \best{$\text{92.0}_{\pm \text{0.9}}$} & $\text{91.4}_{\pm \text{1.0}}$ & $\text{91.3}_{\pm \text{0.9}}$ & $\text{91.0}_{\pm \text{1.0}}$ & $\text{91.6}_{\pm \text{0.8}}$ \\
        \texttt{aloi} & 43.4 & $\text{80.4}_{\pm \text{0.3}}$ & \best{$\text{83.7}_{\pm \text{0.2}}$} & \secondbest{$\text{83.6}_{\pm \text{0.2}}$} & $\text{83.0}_{\pm \text{0.1}}$ &$\text{82.4}_{\pm \text{0.2}}$ &  $\text{80.4}_{\pm \text{0.3}}$ \\
        \texttt{dtd} & 76.8 & $\text{53.2}_{\pm \text{1.0}}$ & \best{$\text{55.8}_{\pm \text{1.5}}$} & $\text{55.1}_{\pm \text{0.9}}$ & \secondbest{$\text{55.4}_{\pm \text{1.1}}$} & $\text{53.8}_{\pm \text{1.0}}$ & $\text{53.2}_{\pm \text{1.0}}$  \\
        \texttt{agnews} & 56.8 & $\text{82.4}_{\pm \text{3.2}}$ & $\text{86.1}_{\pm \text{1.4}}$ & $\text{87.0}_{\pm \text{0.7}}$ & \secondbest{$\text{87.4}_{\pm \text{0.8}}$} & \best{$\text{87.4}_{\pm \text{0.6}}$} & $\text{82.4}_{\pm \text{3.2}}$  \\
        \bottomrule
    \end{tabular}
\end{table}

\textbf{Generalization effect:} Table~\ref{tab:ablation} shows the generalization effect and robustness regarding the hyperparameter $\alpha$, used for sampling the mixing coefficient $\lambda$ in Eq.~\eqref{eq:coefficient}. A key observation is that for the vast majority of tested $\alpha$ values and datasets, integrating our \texttt{mixup} extension into the training of \texttt{annot-mix} improves its generalization performance. Consequently, these performance gains are also robust to some extent regarding the choice of the hyperparameter~$\alpha$. Still, there are a few noteworthy differences, e.g., larger $\alpha$ values tend to perform better when training randomly initialized ResNets, whereas smaller (non-zero) $\alpha$ values tend to provide larger gains when training pretrained models and MLPs. Despite these differences, we fix $\alpha=1$ for all subsequent empirical evaluations. On the one hand, this parameterization corresponds to the choice of a uniform distribution over the mixing coefficient. On the other hand, choosing such a default value allows for a fair comparison with the other multi-annotator classification approaches.

\textbf{Effect of mixing triples with different instances:} Inspired by the idea of Zhang et al.~\cite{zhang2022identifying}, we modify our \texttt{mixup} extension to only combine two triples $(\boldsymbol{x}_n, \boldsymbol{a}_m, \boldsymbol{z}_{nm}), (\boldsymbol{x}_{\hat{n}}, \boldsymbol{a}_{\hat{m}}, \boldsymbol{z}_{{\hat{n}}{\hat{m}}}) \in \mathcal{M}$, if and only if both instances are equal, i.e., $\boldsymbol{x}_n = \boldsymbol{x}_{\hat{n}}$. Evaluating this modified \texttt{mixup} extension while keeping the rest of \texttt{annot-mix} unchanged allows us to study the benefit of mixing triples containing different instances. This modification of \texttt{annot-mix} is not equivalent to the approach of Zhang et al.~\cite{zhang2022identifying}, and we refer to Section~\ref{sec:related-work} for further differences. Table~\ref{tab:ablation} presents the results of \texttt{annot-mix} with this modification in its last column. For the datasets with one class label per instance, the results are identical to the training w/o \texttt{mixup}, as mixing only happens if multiple class labels per instance are available. The performance gains compared to training w/o \texttt{mixup} are noteworthy for the datasets \texttt{letter} and \texttt{flowers102}, while no substantial gains are observed for the other datasets. Furthermore, the performance results across almost all tested datasets fall short of those achieved by our original \texttt{mixup} extension as part of \texttt{annot-mix}. This observation highlights the importance and broader applicability of mixing triples containing different instances.

\subsection{Benchmark Study}

\textbf{Classification models:} Table~\ref{tab:comparative-gt-acc} presents the results for comparing the classification models' performances trained by the different approaches per dataset. As expected, training with the true class labels, i.e., \texttt{true-base}, leads to the best results across all datasets. Comparing the performances of this upper baseline to \texttt{mv-base} as the lower baseline, we clearly observe the negative impact of the noisy class labels. For example, the performance gap between the lower and upper baseline is about \si{40}{\%} for the dataset \texttt{dtd}, which contains the highest fraction of false class labels. The approach \texttt{mv-mixup} strongly reduces this performance gap for almost all datasets and thus confirms the benefit of vanilla \texttt{mixup} in combination with two-stage approaches. For the dataset with multiple class labels per instance, training with \texttt{ds-mixup} yields additional performance gains, establishing it as a strong two-stage competitor to the one-stage approaches. If we now also consider the results of these one-stage approaches, we recognize that \texttt{annot-mix} is the only approach outperforming the two-stage approaches for almost each dataset. The other one-stage approaches often perform inferiorly on datasets with many classes. For example, except \texttt{annot-mix}, all one-stage approaches are worse than \texttt{mv-mixup} for the dataset \texttt{aloi} with $C=1{,}000$ classes. Further, \texttt{annot-mix} outperforms its competitors by considerable margins for four of the five datasets annotated by humans. Comparing the classification models' performances after the last epoch and after the best epoch, selected via a validation set during training, we observe inconsistent improvements in this model selection. This is mainly due to the small size of the validation sets. However, this also shows another advantage of \texttt{annot-mix}, which performs better than its competitors on ten out of eleven datasets when no expensive validation sets with clean labels are available. 

\begin{figure}[!t]
    \centering
    \begin{tikzpicture}
        \node(rankings) at (0, 0) {\rotatebox{90}{\includegraphics[width=0.076\textwidth]{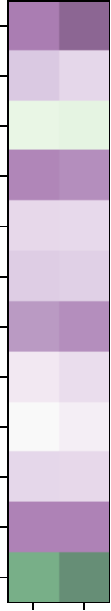}}};

        \node(best-epoch)[above right = -1.32cm and - 0.2cm of rankings] {\rotatebox{60}{\scriptsize Best Epoch}};
        \node(last-epoch)[above right = -0.7cm and -0.2cm of rankings] {\rotatebox{60}{\scriptsize Last Epoch}};

        \node[below left = -0.15cm and -0.65cm of rankings] {\rotatebox{-90}{\scriptsize\texttt{mv-base}}};
        \node[above left = -0.58cm and -0.77cm of rankings] {\scriptsize9.36};
        \node[above left = -0.75cm and -0.62cm of rankings] {\scriptsize $\star$};
        \node[above left = -1.19cm and -0.77cm of rankings] {\scriptsize8.59};
        \node[above left = -1.36cm and -0.62cm of rankings] {\scriptsize $\star$};

        \node[below left = -0.15cm and -1.27cm of rankings] {\rotatebox{-90}{\scriptsize\texttt{mv-mixup}}};
        \node[above left = -0.58cm and -1.39cm of rankings] {\scriptsize6.50};
        \node[above left = -0.75cm and -1.24cm of rankings] {\scriptsize $\star$};
        \node[above left = -1.19cm and -1.39cm of rankings] {\scriptsize6.91};
        \node[above left = -1.36cm and -1.24cm of rankings] {\scriptsize $\star$};

        \node[below left = -0.15cm and -1.89cm of rankings] {\rotatebox{-90}{\scriptsize\texttt{ds-mixup}}};
        \node[above left = -0.58cm and -2.01cm of rankings] {\scriptsize4.41};
        \node[above left = -0.75cm and -1.86cm of rankings] {\scriptsize $\star$};
        \node[above left = -1.19cm and -2.01cm of rankings] {\scriptsize4.55};
        \node[above left = -1.36cm and -1.86cm of rankings] {\scriptsize $\star$};

        \node[below left = -0.15cm and -2.51cm of rankings] {\rotatebox{-90}{\scriptsize\texttt{crowd-layer}}};
        \node[above left = -0.58cm and -2.63cm of rankings] {\scriptsize8.23};
        \node[above left = -0.75cm and -2.48cm of rankings] {\scriptsize $\star$};
        \node[above left = -1.19cm and -2.63cm of rankings] {\scriptsize8.41};
        \node[above left = -1.36cm and -2.48cm of rankings] {\scriptsize $\star$};

        \node[below left = -0.15cm and -3.13cm of rankings] {\rotatebox{-90}{\scriptsize\texttt{trace-reg}}};
        \node[above left = -0.58cm and -3.25cm of rankings] {\scriptsize6.41};
        \node[above left = -0.75cm and -3.10cm of rankings] {\scriptsize $\star$};
        \node[above left = -1.19cm and -3.25cm of rankings] {\scriptsize6.45};
        \node[above left = -1.36cm and -3.10cm of rankings] {\scriptsize $\star$};

        \node[below left = -0.15cm and -3.75cm of rankings] {\rotatebox{-90}{\scriptsize\texttt{conal}}};
        \node[above left = -0.58cm and -3.87cm of rankings] {\scriptsize6.68};
        \node[above left = -0.75cm and -3.72cm of rankings] {\scriptsize $\star$};
        \node[above left = -1.19cm and -3.87cm of rankings] {\scriptsize6.77};
        \node[above left = -1.36cm and -3.72cm of rankings] {\scriptsize $\star$};

        \node[below left = -0.15cm and -4.37cm of rankings] {\rotatebox{-90}{\scriptsize\texttt{union-net}}};
        \node[above left = -0.58cm and -4.49cm of rankings] {\scriptsize8.23};
        \node[above left = -0.75cm and -4.34cm of rankings] {\scriptsize $\star$};
        \node[above left = -1.19cm and -4.49cm of rankings] {\scriptsize8.00};
        \node[above left = -1.36cm and -4.34cm of rankings] {\scriptsize $\star$};

        \node[below left = -0.15cm and -4.98cm of rankings] {\rotatebox{-90}{\scriptsize\texttt{madl}}};
        \node[above left = -0.58cm and -5.10cm of rankings] {\scriptsize6.32};
        \node[above left = -0.75cm and -4.95cm of rankings] {\scriptsize $\star$};
        \node[above left = -1.19cm and -5.10cm of rankings] {\scriptsize5.95};
        \node[above left = -1.36cm and -4.95cm of rankings] {\scriptsize $\star$};

        \node[below left = -0.15cm and -5.60cm of rankings] {\rotatebox{-90}{\scriptsize\texttt{geo-reg-f}}};
        \node[above left = -0.58cm and -5.72cm of rankings] {\scriptsize5.73};
        \node[above left = -0.75cm and -5.57cm of rankings] {\scriptsize $\star$};
        \node[above left = -1.19cm and -5.72cm of rankings] {\scriptsize5.32};
        \node[above left = -1.36cm and -5.57cm of rankings] {\scriptsize $\star$};

        \node[below left = -0.15cm and -6.22cm of rankings] {\rotatebox{-90}{\scriptsize\texttt{geo-reg-w}}};
        \node[above left = -0.58cm and -6.34cm of rankings] {\scriptsize6.45};
        \node[above left = -0.75cm and -6.19cm of rankings] {\scriptsize $\star$};
        \node[above left = -1.19cm and -6.34cm of rankings] {\scriptsize6.50};
        \node[above left = -1.36cm and -6.19cm of rankings] {\scriptsize $\star$};

        \node[below left = -0.15cm and -6.84cm of rankings] {\rotatebox{-90}{\scriptsize\texttt{crowd-ar}}};
        \node[above left = -0.58cm and -6.96cm of rankings] {\scriptsize8.45};
        \node[above left = -0.75cm and -6.81cm of rankings] {\scriptsize $\star$};
        \node[above left = -1.19cm and -6.96cm of rankings] {\scriptsize8.45};
        \node[above left = -1.36cm and -6.81cm of rankings] {\scriptsize $\star$};

        \node[below left = -0.15cm and -7.46cm of rankings] {\rotatebox{-90}{\scriptsize\texttt{annot-mix}}};
        \node[above left = -0.58cm and -7.58cm of rankings] {\scriptsize1.23};
        \node[above left = -0.75cm and -7.43cm of rankings] {\scriptsize};
        \node[above left = -1.19cm and -7.58cm of rankings] {\scriptsize2.09};
        \node[above left = -1.36cm and -7.43cm of rankings] {\scriptsize};
    \end{tikzpicture}
    \caption{Benchmark study: Numbers refer to the approaches' mean ranks across the datasets in Table~\ref{tab:comparative-gt-acc}. Lower values mean better ranks. A star ($\star$) marks that \texttt{annot-mix} performs significantly superior to its respective competitor.}
    \label{fig:rankings}
\end{figure}

For a more compact presentation of the results in Table~\ref{tab:comparative-gt-acc}, we further compute each approach's rank per dataset and report their means in Fig.~\ref{fig:rankings}. Moreover, we evaluate statistical significance at the $0.05$ level by following a common test protocol~\cite{demvsar2006statistical}. Concretely, we perform a Friedman test~\cite{friedman1937use} as an omnibus test with the null hypothesis that all approaches perform the same and observed performance differences are due to randomness. If this null hypothesis is rejected, we proceed with Dunn's post-hoc test~\cite{dunn1961multiple} for pairwise multiple comparisons between \texttt{annot-mix} and each of its competitors. Thereby, we employ Holm's step-down procedure~\cite{holm1979simple} to control for the family-wise error rate. This test protocol is applied to the classification model's performances after the last and the best epoch. The results demonstrate that \texttt{annot-mix} significantly outperforms each competitor. Additionally, it is noteworthy that \texttt{ds-mixup} demonstrates greater robustness across datasets as a two-stage approach compared to its one-stage competitors (with the exception of \texttt{annot-mix}).

\begin{figure}[!t]
    \centering
    \begin{tikzpicture}
        \node(x-axis) at (0.8, 0) {\rotatebox{90}{\scriptsize \texttt{clf-acc} [\%]}};
        \node[right = 4.0cm of x-axis] {\rotatebox{90}{\scriptsize \texttt{clf-acc} [\%]}};
        \node(learning-curves)[above right = -4.45cm and -0.2cm of x-axis] {\includegraphics[width=0.45\textwidth]{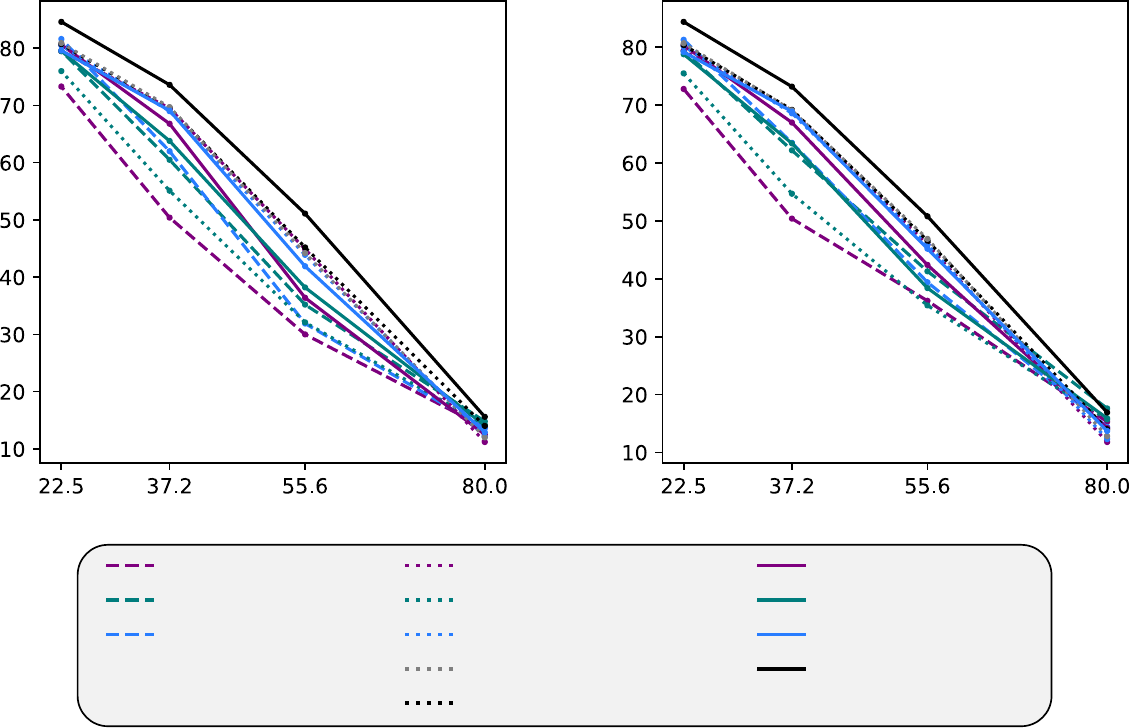}};
        \node[below left = -1.82cm and -3.5cm of learning-curves] {\scriptsize False Label Fraction [\%]};
        \node[below right = -1.82cm and -3.1cm of learning-curves] {\scriptsize False Label Fraction [\%]};
        \node(title)[above left =-0.13cm and -2.6cm of learning-curves] {\scriptsize Last Epoch};
        \node(title)[above right =-0.13cm and -2.3cm of learning-curves] {\scriptsize Best Epoch};
        \node at (3.55, -2.3) {\parbox{2.7cm}{\scriptsize \texttt{mv-base}}};
        \node at (3.55, -2.57) {\parbox{2.7cm}{\scriptsize \texttt{mv-mixup}}};
        \node at (3.55, -2.82) {\parbox{2.7cm}{\scriptsize \texttt{ds-mixup}}};
        \node at (5.7, -2.3) {\parbox{2.7cm}{\scriptsize \texttt{crowd-layer}}};
        \node at (5.7, -2.57) {\parbox{2.7cm}{\scriptsize \texttt{trace-reg}}};
        \node at (5.7, -2.79) {\parbox{2.7cm}{\scriptsize \texttt{union-net}}};
        \node at (5.7, -3.07) {\parbox{2.7cm}{\scriptsize \texttt{geo-reg-w}}};
        \node at (5.7, -3.30) {\parbox{2.7cm}{\scriptsize \texttt{geo-reg-f}}};
        \node at (7.86, -2.29) {\parbox{2.0cm}{\scriptsize \texttt{conal}}};
        \node at (7.86, -2.55) {\parbox{2.0cm}{\scriptsize \texttt{madl}}};
        \node at (7.86, -2.80) {\parbox{2.0cm}{\scriptsize \texttt{crowd-ar}}};
        \node at (7.86, -3.04) {\parbox{2.0cm}{\scriptsize \texttt{annot-mix}}};
    \end{tikzpicture}
    \caption{Benchmark study: Each curve shows the \texttt{clf-acc} [\%] (cf.~Eq.~\eqref{eq:clf-acc}) of an approach across four false label fractions [\%] for \texttt{cifar10h}.}
    \label{fig:noise-rates}
\end{figure}

Figure~\ref{fig:noise-rates} concludes our analysis by showing the approaches' test accuracies for four different variants of the \texttt{cifar10h} dataset as curves. These variants include only class labels from the original dataset's $M=10$, $M=20$, $M=40$, and $M=100$ least accurate annotators, inducing four different fractions of false class labels. As expected, each approach's test accuracy decreases with an increasing false label fraction. Further, the lower baseline \texttt{mv-base} performs worst, as shown by the low-lying dashed purplish curve, whereas the solid black curve of \texttt{annot-mix} mostly remains above the other curves, demonstrating its superior performance across various false label fractions.

\begin{table*}[!ht]
    \setlength{\tabcolsep}{5.0pt}
    \def\arraystretch{1.215}
    \scriptsize
    \caption{
        Benchmark study: \best{Best}, \secondbest{second best}, and \worse{worse than \texttt{mv-base}} results of the \texttt{clf-acc} [\%] (cf.~Eq.~\eqref{eq:clf-acc}) are marked per dataset (column-wise) while \excluded{excluding} results of \texttt{true-base}. Numbers below datasets (second row) show false label fractions [\%]. The results of \texttt{mv-mixup} and \texttt{ds-mixup} are identical for the datasets with only one class label per instance since the Dawid-Skene algorithm~\cite{dawid1979maximum} reduces to majority voting in these cases.
    }
    \label{tab:comparative-gt-acc}
    \centering
    \begin{tabular}{|r|ccccc|cccccc|}
        \toprule
        \multicolumn{1}{|c|}{\multirow{2}{*}{\textBF{\texttt{clf-acc} [\%]}}} & \texttt{mgc} & \texttt{labelme} & \texttt{cifar10h} & \texttt{cifar10n} & \texttt{cifar100n} & \texttt{letter} & \texttt{flowers102} & \texttt{trec6} & \texttt{aloi} & \texttt{dtd} & \texttt{agnews} \\
        \cmidrule(lr){2-2} \cmidrule(lr){3-3} \cmidrule(lr){4-4} \cmidrule(lr){5-5} \cmidrule(lr){6-6} \cmidrule(lr){7-7} \cmidrule(lr){8-8} \cmidrule(lr){9-9} \cmidrule(lr){10-10} \cmidrule(lr){11-11} \cmidrule(lr){12-12}
        & 44.0 & 26.0 & 22.5 & 40.2 & 40.2 & 51.9 & 67.5 & 36.9 & 43.4 & 76.8 & 56.8 \\
        \hline
        \multicolumn{12}{|c|}{\cellcolor{datasetcolor!10} Last Epoch}\\
        \hline
        \texttt{true-base}    & \excluded{$\text{79.6}_{\pm \text{0.9}}$} & \excluded{$\text{93.9}_{\pm \text{0.3}}$} & \excluded{$\text{85.2}_{\pm \text{0.2}}$} & \excluded{$\text{94.0}_{\pm \text{0.2}}$} & \excluded{$\text{74.5}_{\pm \text{0.3}}$} & \excluded{$\text{98.0}_{\pm \text{0.1}}$} & \excluded{$\text{99.5}_{\pm \text{0.0}}$} & \excluded{$\text{93.7}_{\pm \text{0.6}}$} & \excluded{$\text{95.7}_{\pm \text{0.1}}$} & \excluded{$\text{78.1}_{\pm \text{0.3}}$} & \excluded{$\text{92.9}_{\pm \text{0.1}}$\phantom{0}} \\
        \texttt{mv-base}    & $\text{66.1}_{\pm \text{2.8}}$ & $\text{80.5}_{\pm \text{1.1}}$ & $\text{73.3}_{\pm \text{0.2}}$ & $\text{63.8}_{\pm \text{0.6}}$ & $\text{51.9}_{\pm \text{0.3}}$ & $\text{75.6}_{\pm \text{0.7}}$ & $\text{68.0}_{\pm \text{1.4}}$ & $\text{86.1}_{\pm \text{1.1}}$ & $\text{71.9}_{\pm \text{0.3}}$   & $\text{35.6}_{\pm \text{0.9}}$ & $\text{74.3}_{\pm \text{0.5}}$\phantom{0} \\
        \hline
        \texttt{mv-mixup}     & $\text{68.2}_{\pm \text{2.1}}$ & $\text{82.8}_{\pm \text{0.7}}$ & $\text{79.5}_{\pm \text{0.3}}$ & \multirow{2}{*}{$\text{81.3}_{\pm \text{0.4}}$} & \multirow{2}{*}{\secondbest{$\text{60.0}_{\pm \text{0.3}}$}} & $\text{81.3}_{\pm \text{0.4}}$ & $\text{71.7}_{\pm \text{1.3}}$ & $\text{89.0}_{\pm \text{0.9}}$ & \multirow{2}{*}{\secondbest{$\text{81.3}_{\pm \text{0.2}}$}}  & \multirow{2}{*}{$\text{43.0}_{\pm \text{1.1}}$} & \multirow{2}{*}{$\text{74.4}_{\pm \text{0.7}\phantom{0}}$} \\
        \texttt{ds-mixup} &  $\text{68.9}_{\pm \text{1.5}}$ &  $\text{85.3}_{\pm \text{0.4}}$ &  \secondbest{$\text{81.6}_{\pm \text{0.2}}$} & & &  \secondbest{$\text{83.9}_{\pm \text{0.4}}$} &  $\text{78.9}_{\pm \text{0.3}}$ &  $\text{91.0}_{\pm \text{0.5}}$ & & & \\
        \hline
        \texttt{crowd-layer}         & $\text{69.8}_{\pm \text{1.0}}$ & \secondbest{$\text{85.7}_{\pm \text{0.5}}$} & $\text{79.5}_{\pm \text{0.5}}$ & $\text{77.9}_{\pm \text{0.4}}$ & \phantom{0}$\text{\worse{4.8}}_{\pm \text{0.6}}$ & $\text{\worse{56.8}}_{\pm \text{2.3}}$ & $\text{\worse{36.8}}_{\pm \text{2.2}}$ & $\text{91.1}_{\pm\text{0.6}}$ & $\text{\phantom{0}\underline{0.9}}_{\pm \text{0.3}}$  &  $\text{\worse{31.3}}_{\pm \text{1.1}}$ & $\text{85.6}_{\pm \text{0.3\phantom{0}}}$ \\
        \texttt{trace-reg}       & $\text{66.4}_{\pm \text{1.4}}$ & $\text{82.6}_{\pm \text{0.6}}$ & $\text{76.0}_{\pm \text{0.4}}$ & $\text{65.1}_{\pm \text{0.5}}$ & $\text{53.7}_{\pm \text{0.5}}$ & $\text{82.7}_{\pm \text{0.3}}$ & $\text{76.2}_{\pm \text{0.5}}$ & \secondbest{$\text{92.0}_{\pm \text{0.5}}$} & $\text{\worse{60.4}}_{\pm \text{0.5}}$  & $\text{36.6}_{\pm \text{0.5}}$ & \best{$\text{86.7}_{\pm \text{0.2\phantom{0}}}$} \\
        \texttt{conal}      & $\text{69.0}_{\pm \text{0.9}}$ & $\text{83.7}_{\pm \text{0.4}}$ & $\text{80.6}_{\pm \text{0.2}}$ & $\text{77.9}_{\pm \text{0.4}}$ & $\text{\worse{27.4}}_{\pm \text{1.4}}$ & $\text{82.5}_{\pm \text{1.4}}$ & $\text{\worse{52.8}}_{\pm \text{3.1}}$ & $\text{90.1}_{\pm \text{0.7}}$ & $\text{\worse{21.3}}_{\pm \text{1.4}}$  & $\text{40.9}_{\pm \text{1.8}}$ & $\text{76.1}_{\pm \text{0.4\phantom{0}}}$ \\
        \texttt{union-net}  & $\text{68.6}_{\pm \text{1.0}}$ & $\text{85.2}_{\pm \text{0.3}}$ & $\text{80.5}_{\pm \text{0.4}}$ & \secondbest{$\text{81.4}_{\pm \text{0.5}}$} & \phantom{0}$\worse{\text{1.3}}_{\pm \text{0.6}}$ & $\worse{\text{66.3}}_{\pm \text{2.0}}$ & $\worse{\text{43.1}}_{\pm \text{3.0}}$ & $\text{90.1}_{\pm \text{0.3}}$  & $\text{\phantom{0}\worse{0.9}}_{\pm \text{0.2}}$  & $\worse{\text{30.9}}_{\pm \text{1.8}}$ & $\text{86.2}_{\pm \text{0.3\phantom{0}}}$ \\
        \texttt{madl}       & \secondbest{$\text{72.0}_{\pm \text{2.0}}$} & $\text{82.5}_{\pm \text{0.8}}$ & $\text{79.5}_{\pm \text{0.5}}$ & $\text{76.9}_{\pm \text{0.4}}$ & $\worse{\text{42.8}}_{\pm \text{7.4}}$ & $\worse{\text{69.1}}_{\pm \text{4.6}}$ & \secondbest{$\text{85.0}_{\pm \text{1.9}}$} & $\text{91.1}_{\pm \text{0.7}}$ &  $\text{79.2}_{\pm \text{0.3}}$  & \secondbest{$\text{47.2}_{\pm \text{1.8}}$} & $\text{76.0}_{\pm \text{10.8}}$ \\
        \texttt{geo-reg-f}       & $\text{70.2}_{\pm \text{0.7}}$ & $\text{85.4}_{\pm \text{0.5}}$  & $\text{80.7}_{\pm \text{0.4}}$  & $\text{80.5}_{\pm \text{0.3}}$  & $\text{\phantom{0}\worse{8.1}}_{\pm \text{0.9}}$  & $\text{82.4}_{\pm \text{1.9}}$  & $\text{\worse{44.5}}_{\pm \text{3.3}}$  & $\text{91.8}_{\pm \text{0.3}}$  & $\text{\phantom{0}\worse{1.6}}_{\pm \text{0.3}}$   & $\text{\worse{35.2}}_{\pm \text{1.5}}$ & \secondbest{$\text{86.7}_{\pm \text{0.2\phantom{0}}}$} \\
        \texttt{geo-reg-w}       & $\text{70.1}_{\pm \text{0.9}}$ & $\text{85.5}_{\pm \text{0.5}}$  & $\text{80.9}_{\pm \text{0.4}}$  & $\text{79.8}_{\pm \text{0.2}}$  & $\text{\phantom{0}\worse{8.1}}_{\pm \text{0.9}}$  & $\worse{\text{73.9}}_{\pm \text{3.0}}$  & $\text{\worse{44.5}}_{\pm \text{3.3}}$  & $\text{91.9}_{\pm \text{0.6}}$ & $\text{\phantom{0}\worse{1.7}}_{\pm \text{0.3}}$ & $\text{\worse{34.6}}_{\pm \text{1.3}}$ & $\text{82.1}_{\pm \text{9.0\phantom{0}}}$ \\
        \texttt{crowd-ar}   & $\text{69.0}_{\pm \text{1.8}}$ & $\text{84.6}_{\pm \text{0.7}}$ & $\text{79.6}_{\pm \text{0.2}}$ & $\text{80.5}_{\pm \text{0.5}}$ & \phantom{0}$\text{\worse{1.0}}_{\pm \text{0.0}}$ & $\text{78.1}_{\pm \text{2.2}}$ & $\text{\worse{48.2}}_{\pm \text{2.5}}$ & $\text{89.4}_{\pm \text{0.4}}$ & $\text{\phantom{0}\worse{0.1}}_{\pm \text{0.1}}$ & $\text{39.3}_{\pm \text{1.5}}$ & $\text{\worse{72.4}}_{\pm \text{4.8\phantom{0}}}$ \\
        \hline
        \texttt{annot-mix}  & \best{$\text{73.8}_{\pm \text{1.1}}$} & \best{$\text{85.8}_{\pm \text{0.6}}$} & \best{$\text{84.6}_{\pm \text{0.2}}$} & \best{$\text{82.4}_{\pm \text{0.5}}$} & \best{$\text{64.7}_{\pm \text{0.4}}$} & \best{$\text{85.1}_{\pm \text{1.3}}$} & \best{$\text{90.1}_{\pm \text{0.9}}$} & \best{$\text{92.3}_{\pm \text{0.6}}$} & \best{$\text{83.6}_{\pm \text{0.1}}$} & \best{$\text{55.2}_{\pm \text{1.1}}$} & $\text{86.2}_{\pm \text{1.2\phantom{0}}}$ \\
        \hline
        \multicolumn{12}{|c|}{\cellcolor{datasetcolor!10} Best Epoch}\\
        \hline
        
        \texttt{true-base} & \excluded{$\text{78.9}_{\pm \text{1.1}}$} & \excluded{$\text{93.8}_{\pm \text{0.5}}$} & \excluded{$\text{84.8}_{\pm \text{0.5}}$} & \excluded{$\text{93.5}_{\pm \text{0.7}}$} & \excluded{$\text{74.4}_{\pm \text{0.3}}$} & \excluded{$\text{97.6}_{\pm \text{0.6}}$} & \excluded{$\text{99.4}_{\pm \text{0.1}}$} & \excluded{$\text{93.3}_{\pm \text{0.6}}$} & \excluded{$\text{95.7}_{\pm \text{0.1}}$} & \excluded{$\text{77.6}_{\pm \text{0.5}}$} & \excluded{$\text{92.8}_{\pm \text{0.2\phantom{0}}}$} \\
        
        \texttt{mv-base} & $\text{66.8}_{\pm \text{2.6}}$ & $\text{85.5}_{\pm \text{0.8}}$ & $\text{72.8}_{\pm \text{0.8}}$ & $\text{79.1}_{\pm \text{0.7}}$ & $\text{53.2}_{\pm \text{1.1}}$ & $\text{78.4}_{\pm \text{0.9}}$ & $\text{71.5}_{\pm \text{1.2}}$ & $\text{87.4}_{\pm \text{0.6}}$ & $\text{74.9}_{\pm \text{0.4}}$ & $\text{46.9}_{\pm \text{0.9}}$ & $\text{77.1}_{\pm \text{0.8\phantom{0}}}$ \\

        \hline
        
        \texttt{mv-mixup} & $\text{67.5}_{\pm \text{3.3}}$ & $\worse{\text{85.1}}_{\pm \text{1.2}}$ & $\text{79.4}_{\pm \text{0.4}}$ & \multirow{2}{*}{\secondbest{$\text{82.8}_{\pm \text{0.6}}$}} & \multirow{2}{*}{\secondbest{$\text{60.2}_{\pm \text{0.4}}$}} & $\text{81.3}_{\pm \text{0.7}}$ & $\text{71.8}_{\pm \text{1.3}}$ & $\text{88.2}_{\pm \text{1.0}}$ & \multirow{2}{*}{\secondbest{$\text{81.2}_{\pm \text{0.3}}$}} & \multirow{2}{*}{$\text{\worse{46.2}}_{\pm \text{1.4}}$} & \multirow{2}{*}{$\worse{\text{76.6}}_{\pm \text{1.4\phantom{0}}}$} \\

        \texttt{ds-mixup} & $\text{69.0}_{\pm \text{2.0}}$ & $\text{87.2}_{\pm \text{1.2}}$ & \secondbest{$\text{81.3}_{\pm \text{0.5}}$} & & & \secondbest{$\text{83.4}_{\pm \text{0.5}}$} & $\text{79.5}_{\pm \text{0.6}}$ & $\text{90.8}_{\pm \text{0.9}}$ & & &  \\
        
        \hline
        \texttt{crowd-layer} & $\text{69.0}_{\pm \text{1.1}}$ & $\text{87.3}_{\pm \text{0.5}}$ & $\text{79.4}_{\pm \text{0.5}}$ & $\text{81.9}_{\pm \text{0.5}}$ & \phantom{0}$\text{\worse{4.5}}_{\pm \text{0.5}}$ & $\text{\worse{57.2}}_{\pm \text{2.1}}$ & $\text{\worse{36.9}}_{\pm \text{2.5}}$ & $\text{90.7}_{\pm \text{0.8}}$ & $\text{\phantom{0}\worse{1.0}}_{\pm \text{0.3}}$ & $\text{\worse{31.9}}_{\pm \text{1.2}}$ & $\text{86.2}_{\pm \text{0.6\phantom{0}}}$ \\
        
        \texttt{trace-reg} & $\text{67.8}_{\pm \text{2.1}}$ & $\text{85.8}_{\pm \text{0.7}}$ & $\text{75.5}_{\pm \text{0.6}}$ & $\text{79.2}_{\pm \text{0.8}}$ & $\text{54.3}_{\pm \text{0.8}}$ & $\text{82.8}_{\pm \text{0.5}}$ & $\text{77.9}_{\pm \text{0.4}}$ & $\text{91.4}_{\pm \text{0.8}}$ & $\text{\worse{64.5}}_{\pm \text{0.9}}$ & $\worse{\text{46.9}}_{\pm \text{1.0}}$ & $\text{86.2}_{\pm \text{0.6\phantom{0}}}$ \\
        
        \texttt{conal} & $\text{69.2}_{\pm \text{2.0}}$ & $\text{87.1}_{\pm \text{0.6}}$ & $\text{80.3}_{\pm \text{0.4}}$ & $\text{79.9}_{\pm \text{0.6}}$ & $\text{\worse{27.1}}_{\pm \text{1.6}}$ & $\text{82.8}_{\pm \text{1.3}}$ & $\text{\worse{52.8}}_{\pm \text{3.0}}$ & $\text{90.0}_{\pm \text{0.4}}$ & $\text{\worse{21.2}}_{\pm \text{1.4}}$ & $\text{\worse{41.8}}_{\pm \text{1.5}}$ & $\text{79.3}_{\pm \text{1.1\phantom{0}}}$ \\
        
        \texttt{union-net} & $\text{68.5}_{\pm \text{1.7}}$ & \best{$\text{87.5}_{\pm \text{0.5}}$} & $\text{80.2}_{\pm \text{0.6}}$ & $\text{82.0}_{\pm \text{0.4}}$ & \phantom{0}$\text{\worse{4.0}}_{\pm \text{0.6}}$ & $\text{\worse{66.8}}_{\pm \text{1.9}}$ & $\text{\worse{44.0}}_{\pm \text{3.3}}$ & $\text{89.8}_{\pm \text{0.6}}$ & $\text{\phantom{0}\worse{0.9}}_{\pm \text{0.2}}$ & $\text{\worse{33.6}}_{\pm \text{1.0}}$ & \secondbest{$\text{87.1}_{\pm \text{0.4\phantom{0}}}$} \\
        
        \texttt{madl} & \secondbest{$\text{72.4}_{\pm \text{1.8}}$} & $\text{86.5}_{\pm \text{0.8}}$ & $\text{78.8}_{\pm \text{1.2}}$ & $\text{80.5}_{\pm \text{0.9}}$ & $\text{\worse{42.7}}_{\pm \text{7.3}}$ & $\text{\worse{71.2}}_{\pm \text{4.0}}$ & \secondbest{$\text{85.1}_{\pm \text{1.9}}$} & $\text{91.1}_{\pm \text{0.8}}$ & $\text{79.0}_{\pm \text{0.4}}$ & \secondbest{$\text{47.7}_{\pm \text{1.3}}$} & $\text{78.0}_{\pm \text{10.0}}$ \\
        
        \texttt{geo-reg-f}       & $\text{70.4}_{\pm \text{0.8}}$ & \secondbest{$\text{87.4}_{\pm \text{0.6}}$}  & $\text{80.4}_{\pm \text{0.7}}$  & $\text{81.9}_{\pm \text{0.5}}$  & $\text{\phantom{0}\worse{7.9}}_{\pm \text{1.0}}$  & $\text{82.4}_{\pm \text{1.5}}$  & $\text{\worse{44.9}}_{\pm \text{3.3}}$  & \secondbest{$\text{91.5}_{\pm \text{0.9}}$} & $\text{\phantom{0}\worse{1.6}}_{\pm \text{0.3}}$ & $\text{\worse{35.9}}_{\pm \text{1.3}}$ & \best{$\text{87.4}_{\pm \text{0.4\phantom{0}}}$} \\
        
        \texttt{geo-reg-w}       & $\text{69.8}_{\pm \text{1.1}}$ & $\text{87.3}_{\pm \text{0.4}}$  & $\text{80.7}_{\pm \text{0.5}}$  & $\text{81.5}_{\pm \text{0.5}}$  & $\text{\phantom{0}\worse{7.9}}_{\pm \text{1.1}}$  & $\worse{\text{74.2}}_{\pm \text{2.4}}$  & $\text{\worse{44.9}}_{\pm \text{3.3}}$  & \best{$\text{91.6}_{\pm \text{1.0}}$} & $\text{\phantom{0}\worse{1.6}}_{\pm \text{0.3}}$ & $\text{\worse{35.8}}_{\pm \text{1.4}}$ & $\text{83.1}_{\pm \text{8.1\phantom{0}}}$  \\
        
        \texttt{crowd-ar} & $\text{70.4}_{\pm \text{2.0}}$ & $\text{87.3}_{\pm \text{0.5}}$ & $\text{79.2}_{\pm \text{0.7}}$ & $\text{80.4}_{\pm \text{0.4}}$ & $\text{\phantom{0}\worse{4.3}}_{\pm \text{0.8}}$ & $\text{\worse{77.8}}_{\pm \text{2.2}}$ & $\text{\worse{48.2}}_{\pm \text{2.4}}$ & $\text{89.1}_{\pm \text{1.0}}$ & $\text{\phantom{0}\worse{1.0}}_{\pm \text{0.4}}$ & $\text{\worse{40.1}}_{\pm \text{1.5}}$ & $\text{\worse{76.3}}_{\pm \text{2.7\phantom{0}}}$\\
        \hline
        \texttt{annot-mix} & \best{$\text{73.8}_{\pm \text{2.1}}$} & $\text{86.5}_{\pm \text{0.7}}$ & \best{$\text{84.4}_{\pm \text{0.3}}$} & \best{$\text{83.2}_{\pm \text{0.8}}$} & \best{$\text{64.1}_{\pm 
        \text{0.6}}$} & \best{$\text{84.9}_{\pm \text{1.6}}$} & \best{$\text{90.0}_{\pm \text{0.9}}$} & $\text{91.4}_{\pm \text{1.0}}$ & \best{$\text{83.5}_{\pm \text{0.1}}$} & \best{$\text{55.1}_{\pm \text{0.9}}$} & $\text{87.0}_{\pm \text{0.7\phantom{0}}}$\\
        \bottomrule
    \end{tabular}
\end{table*}

\begin{table*}[!ht]
    \setlength{\tabcolsep}{5.5pt}
    \def\arraystretch{1.22}
    \scriptsize
    \caption{
        Benchmark study: \best{Best} and \secondbest{second best} results of the \texttt{perf-auroc} [\%] are marked per dataset (row-wise). Numbers to the right of the dataset names (second column) show false label fractions [\%].
    }
    \label{tab:comparative-ap-auroc}
    \centering
    \begin{tabular}{|r|c|cccccccc|c|}
        \toprule
        
        \multicolumn{2}{|c|}{\textBF{\texttt{perf-auroc} [\%]}} & \texttt{crowd-layer} & \texttt{trace-reg} & \texttt{conal} & \texttt{union-net} & \texttt{madl} & \texttt{geo-reg-f} & \texttt{geo-reg-w} & \texttt{crowd-ar} & \texttt{annot-mix} \\
                
        \hline
        
        \multicolumn{11}{|c|}{\cellcolor{datasetcolor!10} Last Epoch}\\
        
        \hline
        
        \texttt{letter} & 51.9 & $\text{78.0}_{\pm \text{1.5}}$ & $\text{87.5}_{\pm \text{0.0}}$ & $\text{63.0}_{\pm \text{0.3}}$ & $\text{86.6}_{\pm \text{1.0}}$ & $\text{84.2}_{\pm \text{2.4}}$ & \secondbest{$\text{91.9}_{\pm \text{0.9}}$} & $\text{87.3}_{\pm \text{1.2}}$ & $\text{61.9}_{\pm \text{0.6}}$ & \best{$\text{93.1}_{\pm \text{0.7}}$}\\
        
        \texttt{flowers102} & 67.5 & $\text{67.2}_{\pm \text{0.6}}$ & $\text{77.8}_{\pm \text{0.8}}$ & $\text{61.0}_{\pm \text{0.4}}$ & $\text{69.5}_{\pm \text{1.0}}$ & \secondbest{$\text{86.0}_{\pm \text{0.8}}$} & $\text{68.7}_{\pm \text{0.9}}$ & $\text{68.7}_{\pm \text{0.9}}$ & $\text{55.4}_{\pm \text{1.1}}$ & \best{$\text{90.4}_{\pm \text{0.3}}$}\\
        
        \texttt{trec6} & 36.9 & \best{$\text{95.5}_{\pm \text{0.1}}$} & $\text{93.8}_{\pm \text{0.1}}$ & $\text{56.3}_{\pm \text{0.2}}$ & \secondbest{$\text{94.7}_{\pm \text{0.1}}$} & $\text{94.6}_{\pm \text{0.3}}$ & $\text{93.9}_{\pm \text{0.1}}$ & $\text{93.8}_{\pm \text{0.1}}$ & $\text{56.3}_{\pm \text{1.4}}$ & $\text{94.2}_{\pm \text{0.1}}$\\
        
        \texttt{aloi} & 43.4 & $\text{48.1}_{\pm \text{1.5}}$ & $\text{70.8}_{\pm \text{0.2}}$ & $\text{52.2}_{\pm \text{0.2}}$ & $\text{56.2}_{\pm \text{1.2}}$ & \secondbest{$\text{85.9}_{\pm \text{0.4}}$} & $\text{55.4}_{\pm \text{1.5}}$ & $\text{55.6}_{\pm \text{1.5}}$ & $\text{50.5}_{\pm \text{4.4}}$ & \best{$\text{88.7}_{\pm \text{0.1}}$}\\
        
        \texttt{dtd} & 76.8 & $\text{65.2}_{\pm \text{1.2}}$ & $\text{55.4}_{\pm \text{0.2}}$ & $\text{58.0}_{\pm \text{0.3}}$ & $\text{65.9}_{\pm \text{1.3}}$ & \secondbest{$\text{75.4}_{\pm \text{1.4}}$} & $\text{65.8}_{\pm \text{0.8}}$ & $\text{66.1}_{\pm \text{0.7}}$ & $\text{56.9}_{\pm \text{0.9}}$ & \best{$\text{81.6}_{\pm \text{0.6}}$}\\
        
        \texttt{agnews} & 56.8 & \best{$\text{93.0}_{\pm \text{0.1}}$} & $\text{92.1}_{\pm \text{0.1}}$ & $\text{61.7}_{\pm \text{0.1}}$ & \secondbest{$\text{93.0}_{\pm \text{0.1}}$} & $\text{88.1}_{\pm \text{7.5}}$ & $\text{92.2}_{\pm \text{0.1}}$ & $\text{89.4}_{\pm \text{5.5}}$ & $\text{63.4}_{\pm \text{1.3}}$ & $\text{92.7}_{\pm \text{0.5}}$\\
        
        \hline
        
        \multicolumn{11}{|c|}{\cellcolor{datasetcolor!10} Best Epoch}\\
        
        \hline
        
        \texttt{letter} & 51.9 & $\text{79.9}_{\pm \text{1.5}}$ & $\text{87.8}_{\pm \text{0.3}}$ & $\text{62.8}_{\pm \text{0.2}}$ & $\text{86.7}_{\pm \text{0.9}}$ & $\text{83.8}_{\pm \text{3.6}}$ & \secondbest{$\text{91.4}_{\pm \text{1.0}}$} & $\text{87.0}_{\pm \text{1.3}}$ & $\text{62.0}_{\pm \text{0.5}}$ & \best{$\text{93.0}_{\pm \text{0.7}}$}\\
        
        \texttt{flowers102} & 67.5 & $\text{65.9}_{\pm \text{3.5}}$ & $\text{78.2}_{\pm \text{0.4}}$ & $\text{60.9}_{\pm \text{0.4}}$ & $\text{69.7}_{\pm \text{1.1}}$ & \secondbest{$\text{85.9}_{\pm \text{0.9}}$} & $\text{69.0}_{\pm \text{0.9}}$ & $\text{69.0}_{\pm \text{0.9}}$ & $\text{54.8}_{\pm \text{1.5}}$ & \best{$\text{90.4}_{\pm \text{0.2}}$}\\
        
        \texttt{trec6} & 36.9 & \best{$\text{95.4}_{\pm \text{0.2}}$} & $\text{93.7}_{\pm \text{0.1}}$ & $\text{56.5}_{\pm \text{0.4}}$ & \secondbest{$\text{94.6}_{\pm \text{0.2}}$} & $\text{94.5}_{\pm \text{0.3}}$ & $\text{93.8}_{\pm \text{0.2}}$ & $\text{93.7}_{\pm \text{0.1}}$ & $\text{56.3}_{\pm \text{1.4}}$ & $\text{94.0}_{\pm \text{0.3}}$\\
        
        \texttt{aloi} & 43.4 & $\text{54.1}_{\pm \text{2.1}}$ & $\text{63.9}_{\pm \text{1.1}}$ & $\text{52.1}_{\pm \text{0.3}}$ & $\text{56.1}_{\pm \text{1.3}}$ & \secondbest{$\text{86.2}_{\pm \text{0.5}}$} & $\text{55.1}_{\pm \text{1.7}}$ & $\text{55.3}_{\pm \text{1.7}}$ & $\text{50.8}_{\pm \text{4.0}}$ & \best{$\text{88.7}_{\pm \text{0.1}}$}\\
        
        \texttt{dtd} & 76.8 & $\text{62.6}_{\pm \text{1.9}}$ & $\text{57.9}_{\pm \text{0.5}}$ & $\text{58.5}_{\pm \text{0.4}}$ & $\text{61.6}_{\pm \text{1.2}}$ & \secondbest{$\text{74.0}_{\pm \text{6.0}}$} & $\text{64.0}_{\pm \text{1.4}}$ & $\text{63.6}_{\pm \text{1.8}}$ & $\text{56.0}_{\pm \text{1.3}}$ & \best{$\text{81.5}_{\pm \text{0.6}}$}\\
        
        \texttt{agnews} & 56.8 & \secondbest{$\text{93.0}_{\pm \text{0.2}}$} & $\text{92.4}_{\pm \text{0.1}}$ & $\text{61.3}_{\pm \text{0.8}}$ & \best{$\text{93.1}_{\pm \text{0.2}}$} & $\text{89.2}_{\pm \text{6.3}}$ & $\text{92.4}_{\pm \text{0.1}}$ & $\text{89.7}_{\pm \text{5.3}}$ & $\text{63.1}_{\pm \text{1.3}}$ & $\text{92.7}_{\pm \text{0.4}}$ \\
        \bottomrule
    \end{tabular}
\end{table*}

\textbf{Annotator models:} Table~\ref{tab:comparative-ap-auroc} presents the results for comparing the annotator models' performances. Intuitively, a high score implies that the corresponding annotator model can accurately predict whether an annotator will provide a correct or false class label for a given instance. We include only the results for the datasets with simulated annotators because the test sets of the other datasets were not annotated by error-prone humans. Further, only the related approaches that train an annotator model are considered. For the results after the last and best epoch, we observe that \texttt{annot-mix} performs best on four of the six datasets while providing competitive results for the other two datasets. As a result, our approach has the potential to be used in applications where it is important to obtain accurate predictions of the annotators' performances, e.g., when selecting the best annotator to provide class labels in an active learning setting~\cite{herde2021survey}.

\section{Conclusion and Outlook}
\label{sec:conclusion}
In this article, we proposed our approach \texttt{annot-mix} addressing the practical challenge of learning from noisy class labels provided by multiple annotators. It maximizes the marginal likelihood of the observed noisy class labels during the joint training of a classification and an annotator model, effectively separating the noise from the true labels. An essential property of our approach is the integration of our novel \texttt{mixup} extension, which convexly combines triples of instances, annotators, and noisy class labels. This data augmentation and regularization technique makes memorizing individual noisy class labels more difficult and thus reduces the risk of overfitting. An extensive empirical evaluation with eleven datasets of three data modalities demonstrated that \texttt{annot-mix} significantly outperforms current state-of-the-art multi-annotator classification approaches.

A future research direction for our work is to adopt ideas of other \texttt{mixup} extensions such as \texttt{cut-mix}~\cite{yun2019cutmix} for image data or \texttt{manifold-mixup}~\cite{verma2019manifold} for combining hidden states of both, the classification and annotator model. Moreover, throughout this article, we used only one-hot encoded representations of the annotators since no datasets with metadata about the annotators were available for our evaluation. Accordingly, collecting such datasets and making them publicly available to the research community would allow us to evaluate the benefit of such metadata. The annotator model's results of \texttt{annot-mix} suggest its potential for querying the most accurate annotators for the most informative instances in active learning settings~\cite{herde2021survey}. A general challenge of multi-annotator classification is reliance on small validation sets, which can render model selection unreliable. Future work could focus on developing model selection processes without necessitating large validation sets~\cite{yuan2024early}. Finally, the extension of \texttt{annot-mix} toward related task types, such as semantic segmentation, by adjusting the classification and annotator model architectures would further broaden its practical use.

\section{Ethical Statement}
\label{sec:ethical-statement}
We confirm that our research refrains from any experimentation with humans. Yet, we emphasize the issue that human annotators, particularly crowdworkers, often endure difficult working conditions~\cite{bhatti2020general}, e.g., minimum job security and low salaries, despite their essential contributions to advancing machine learning research and applications. Although \texttt{annot-mix} allows assessing annotators' performances, we recommend adhering to strict guidelines to avoid unjustified discrimination against annotators. Furthermore, we emphasize our work's empirical nature and, therefore, suggest a thorough empirical evaluation before its application to safety-critical domains.



\begin{ack}
    This work was funded by the ALDeep project through the University of Kassel (grant number: P/681). Moreover, we thank Lukas Rauch for his insightful comments and discussions, which further improved this article. Finally, we acknowledge the usage of the public domain animal images of the jaguar (credit: Hollingsworth, John and Karen, USFWS) and leopard (credit: USFWS) in Figs.~\ref{fig:abstract},~\ref{fig:probabilistic-graphical-model},~\ref{fig:architecture}, and~\ref{fig:mix-up-illustration}.
\end{ack}


\bibliography{references}

\end{document}